\documentclass{article} 
\usepackage{iclr2027_conference,times}

\usepackage{amsmath,amsfonts,bm}

\def\eqref#1{equation~\ref{#1}}

\def\1{\bm{1}}

\DeclareMathAlphabet{\mathsfit}{\encodingdefault}{\sfdefault}{m}{sl}
\SetMathAlphabet{\mathsfit}{bold}{\encodingdefault}{\sfdefault}{bx}{n}

\newcommand{\R}{\mathbb{R}}

\usepackage{hyperref}
\usepackage{url}
\usepackage{graphicx}
\usepackage{placeins}
\usepackage{cleveref}
\usepackage{tikz}
\usepackage{subcaption}
\usepackage{booktabs}
\usepackage{tabularx}
\usepackage{float}

\newcommand{\method}{Tree-VQ}

\title{Tree-VQ: Progressive Image Compression from Pretrained Vector Quantizers}

\author{
Mingming Ma, Xinkun Wang, Tianyi Xu, Qingyu Luo, Fu Li, Yi Niu\thanks{Corresponding author: \texttt{niuyi@mail.xidian.edu.cn}} \\
School of Artificial Intelligence \\
Xidian University \\
\texttt{mamingming@xidian.edu.cn} \\
\texttt{\{25171213993,21009200131,23171110716\}@stu.xidian.edu.cn} \\
\texttt{lifu@mail.xidian.edu.cn}
}
\iclrpreprintcopy

\begin{document}

\maketitle

\begin{abstract}
Progressive image compression requires a single embedded representation whose received prefixes can be decoded without re-encoding the source. Modern vector-quantized (VQ) image models provide strong discrete endpoint representations, but conventional flat codeword indices do not define meaningful intermediate states for a neural decoder. We present \method, a post-hoc conversion of a pretrained flat VQ tokenizer into a fine-grained, arbitrary-prefix progressive representation while preserving its encoder assignments and every learned leaf vector. The key idea is to organize the original codebook into a balanced binary hierarchy, associate explicit representations with internal nodes, and transmit branch decisions in depth-major order. Consequently, once the image header is available, \emph{every payload prefix} uniquely specifies a valid latent state: each additional branch bit refines exactly one token, and transmission can therefore be truncated at essentially any payload position rather than only at a small number of stage boundaries. We further adapt one shared decoder on the complete-depth and mixed-depth latent states encountered under such arbitrary truncation, making these densely spaced prefixes useful for reconstruction rather than merely syntactically decodable. On Kodak, \method~achieves a DISTS-based BD-rate saving of 52.1\% relative to ProGIC, while exposing thousands of valid arbitrary-prefix operating points from a single embedded bitstream.
\end{abstract}

\section{Introduction}
\label{sec:intro}

Progressive transmission has a distinct role at low bitrates. While many existing progressive methods emphasize high-rate refinement or span broad bitrate ranges, stringent bandwidth budgets can force a receiver to stop after a short prefix. A single stream that provides a useful reconstruction early and refines it as more bits arrive directly serves this constraint.
Learned image compression has progressed from nonlinear transform coding and entropy models to discrete tokenizers and perceptual or generative decoders
\citep{balle2017end,balle2018variational,minnen2018joint,mentzer2020high,yang2023lossy}.
Many learned codecs support multiple rate points, but \emph{variable-rate} coding is not necessarily \emph{progressive}: changing the target rate may require re-encoding the image, whereas progressive transmission requires one embedded bitstream whose received prefixes remain decodable as more bits arrive. Classical codecs such as EZW, SPIHT, and JPEG 2000 provide this property by progressively refining a single representation
\citep{shapiro1993embedded,said1996new,skodras2001jpeg}.
Recent learned methods have revisited progressive coding through nested quantization, ordered latents, scalable representations, and hierarchical quantizers
\citep{lee2022dpict,jeon2023ctc,hojjat2023progdtd,zhai2025deepfgs,lee2026deephq,Cao_2026_CVPR}.

\begin{figure}[tbp]
	\centering
	\includegraphics[width=0.96\linewidth]{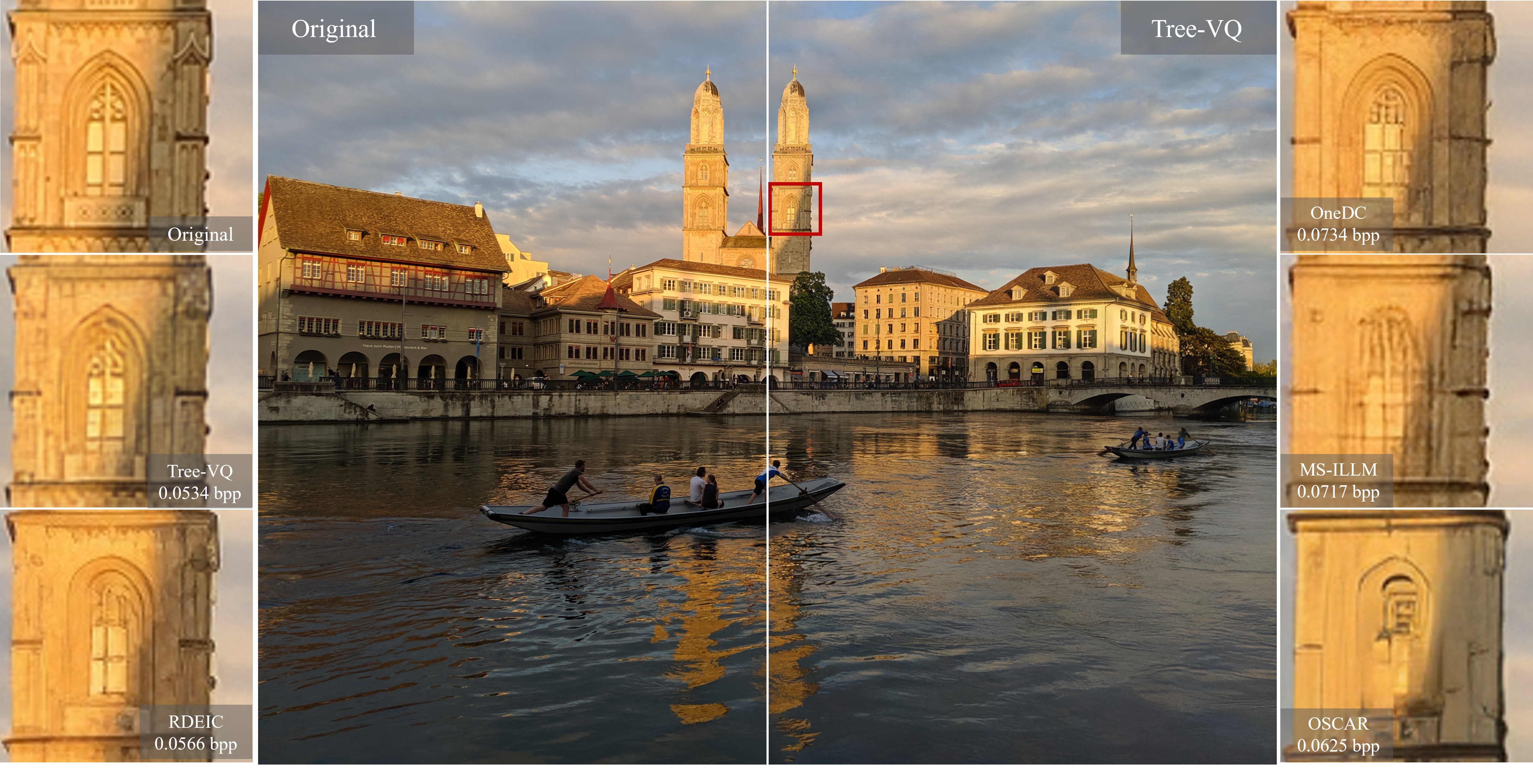}
	\caption{Qualitative cross-method preview on a CLIC2020 test image. The
	central pair shows the original and Tree-VQ reconstruction; side crops show
	OneDC, MS-ILLM, RDEIC, and OSCAR at their displayed rates.}
	\label{fig:tree_recon2}
\end{figure}

Vector quantization (VQ) is attractive for low-rate perceptual compression because it provides compact discrete latent representations
\citep{van2017neural,esser2021taming,lee2022autoregressive,feng2023nvtc}.
However, a conventional flat VQ index is an atomic symbol: truncating its binary label identifies a subset of possible codewords but does not specify a meaningful latent vector for the decoder. As a result, a pretrained flat VQ tokenizer does not naturally support reconstruction from arbitrary partial indices. Existing progressive VQ approaches typically introduce progressivity during model design or training. For example, ProGIC performs progressive refinement through a finite sequence of residual VQ stages \citep{Cao_2026_CVPR}, so its available operating points are tied to the number of quantization stages.

To address this setting, we introduce \method, a codebook-based paradigm for progressive learned image compression that post-hoc augments a pretrained flat VQ tokenizer. We keep the tokenizer's encoder assignments and all learned leaf vectors fixed, and organize those leaves into a balanced binary hierarchy with an explicit representative vector at every internal node. This constraint preserves the endpoint semantics of the pretrained tokenizer while supplying intermediate latent states for progressive transmission. Branch decisions are serialized in depth-major order; every payload prefix then determines one valid ancestor per token, and each additional branch bit refines one token. The intermediate states must also be made compatible with neural synthesis. A decoder trained only on leaf embeddings has never observed these states, so we adapt one shared synthesis transform using complete-depth and mixed-depth latent tensors produced during progressive transmission. The same decoder maps every logical prefix to an image, including prefixes that stop inside a depth plane, and is trained to make these states useful for perceptual rate--distortion reconstruction.

\method~combines dense progressive rate control with strong perceptual rate--distortion performance in the ultra-low-rate regime. On DISTS, Tree-VQ achieves BD-rate savings of $52.1\%$, $49.0\%$, and $53.4\%$ relative to ProGIC on Kodak, DIV2K, and CLIC2020, respectively. Results on DIV2K and CLIC2020 provide additional evidence for this low-rate performance.

Our contributions are:
\begin{itemize}
\item \textbf{A codebook-based paradigm under a fixed pretrained tokenizer.}
	We introduce Tree-VQ, which post-hoc augments a pretrained flat VQ tokenizer with a balanced binary hierarchy and explicit internal-node representatives. The encoder assignments and all learned leaf vectors remain fixed, so the construction adds progressive intermediate states while retaining the tokenizer's endpoint representation.
	
	\item \textbf{Fine-grained intermediate representations for arbitrary prefixes.}
	We construct a single embedded bitstream in which every payload prefix defines a valid latent representation and each additional branch bit refines exactly one token. This decouples progressive granularity from the number of quantization stages and yields thousands of valid truncation points from one encoded representation.
	
\item \textbf{Shared-decoder adaptation for perceptual progressive reconstruction.}
	We adapt one shared decoder to complete-depth and mixed-depth prefix states while keeping the encoder and leaf vocabulary fixed. This makes the intermediate states usable for perceptual rate--distortion reconstruction across the progressive stream, with strong ultra-low-rate results on Kodak, DIV2K, and CLIC2020.
\end{itemize}

\section{Related Work}
\label{sec:related}

\subsection{Progressive and scalable learned image compression}

Learned image compression has evolved from end-to-end transform coding
\citep{balle2017end,theis2017lossy,toderici2017full,rippel2017real}
to hyperprior and context-based entropy models
\citep{balle2018variational,minnen2018joint,cheng2020learned,he2022elic}.
Variable-rate models can operate at multiple quality levels
\citep{choi2019variable},
but changing the target rate does not necessarily yield prefixes of one already transmitted representation.

Progressive learned codecs instead impose nested structure on the transmitted representation.
DPICT and CTC progressively refine trit planes
\citep{lee2022dpict,jeon2023ctc};
ProgDTD learns ordered information with tail-drop training
\citep{hojjat2023progdtd};
DeepFGS and Efficient-PIC introduce scalable enhancement information through learned ordering or masking
\citep{zhai2025deepfgs,presta2025efficient};
and DeepHQ directly learns a hierarchical quantizer
\citep{lee2026deephq}. ARPC combines multi-scale residual quantization with
autoregressive prediction of omitted scales and caption side information
\citep{zhang2026autoregressive}.
Together, these methods expose progressive operating points through latent
groups, planes, masks, learned hierarchies, or autoregressive completion.

Among VQ-based methods, ProGIC uses successive residual-VQ stages for progressive low-rate reconstruction
\citep{Cao_2026_CVPR}, whereas RDVQ-adj obtains adjustable-rate reconstruction by transmitting a partial discrete representation and completing the remaining representation with its learned model
\citep{Jiang_2026_CVPR}.
Both provide meaningful rate control, but their progressive granularity is determined by the structure used during model training or inference.
In contrast, \method~converts an already trained flat VQ vocabulary into a binary hierarchy and explicitly transmits branch decisions.
This yields valid intermediate latent states not only at a small set of stage boundaries but also within each depth plane.

\subsection{Vector quantization and perceptual low-rate compression}

Vector quantization has a long history in lossy compression
\citep{linde1980algorithm}.
Tree-structured VQ organizes prototypes hierarchically and supports coarse-to-fine transmission
\citep{gray1992image},
while progressive index-assignment methods showed that a full-search VQ codebook can also be reorganized for intermediate reconstruction
\citep{riskin1994index}.
Our setting differs in that both the leaf codebook and encoder assignments come from a pretrained neural tokenizer and remain fixed; only the hierarchy and decoder adaptation are introduced post hoc.

Modern neural VQ models such as VQ-VAE, VQ-VAE-2, and VQGAN established discrete learned codebooks as strong image representations
\citep{van2017neural,razavi2019generating,esser2021taming},
with later work improving residual quantization, transform coding, and codebook utilization
\citep{lee2022autoregressive,feng2023nvtc,mentzer2024finite,zhu2024scaling,zhu2025addressing}.
At very low rates, perceptual and generative codecs such as HiFiC, MS-ILLM, PerCo, RDEIC, DLF, OSCAR, OneDC, and StableCodec further improve visual quality
\citep{mentzer2020high,muckley2023improving,careil2024towards,li2025rdeic,xue2025dlf,guo2025oscar,xue2025one,zhang2025stablecodec}.
Tree-VQ complements these advances by adding dense progressive structure to a pretrained discrete representation while preserving its learned leaf vocabulary.

\section{Method}
\label{sec:method}

\subsection{Problem setting}
Let $E_{\theta}$ be a pretrained analysis transform and $G_{\phi_0}$ its synthesis transform. For an image $x\in\R^{H\times W\times 3}$, $E_\theta$ produces $D$-dimensional tokens over spatial strides $s\in\mathcal S$. Concatenating all scales in a fixed order gives
$Z=(z_1,\ldots,z_N)$, where $N=\sum_s HW/s^2$.

A flat codebook $\mathcal C_{\mathrm{leaf}}=\{c_j\}_{j=1}^{K}$ assigns each token to its nearest codeword,
\begin{equation}
	j_i=\arg\min_{j\in\{1,\ldots,K\}}\|z_i-c_j\|_2^2,
	\qquad
	\hat x_{\mathrm{flat}}=G_{\phi_0}(c_{j_1},\ldots,c_{j_N}).
	\label{eq:flatvq}
\end{equation}
We assume $K=2^L$. Our conversion keeps $E_\theta$, the nearest-leaf assignments $j_i$, and every leaf vector $c_j$ fixed, so the full-depth latent representation remains identical to that of the pretrained tokenizer. Decoder adaptation then optimizes a shared synthesis transform for this endpoint and the newly introduced intermediate states.

\begin{figure}[htbp]
	\centering
	\includegraphics[width=\linewidth]{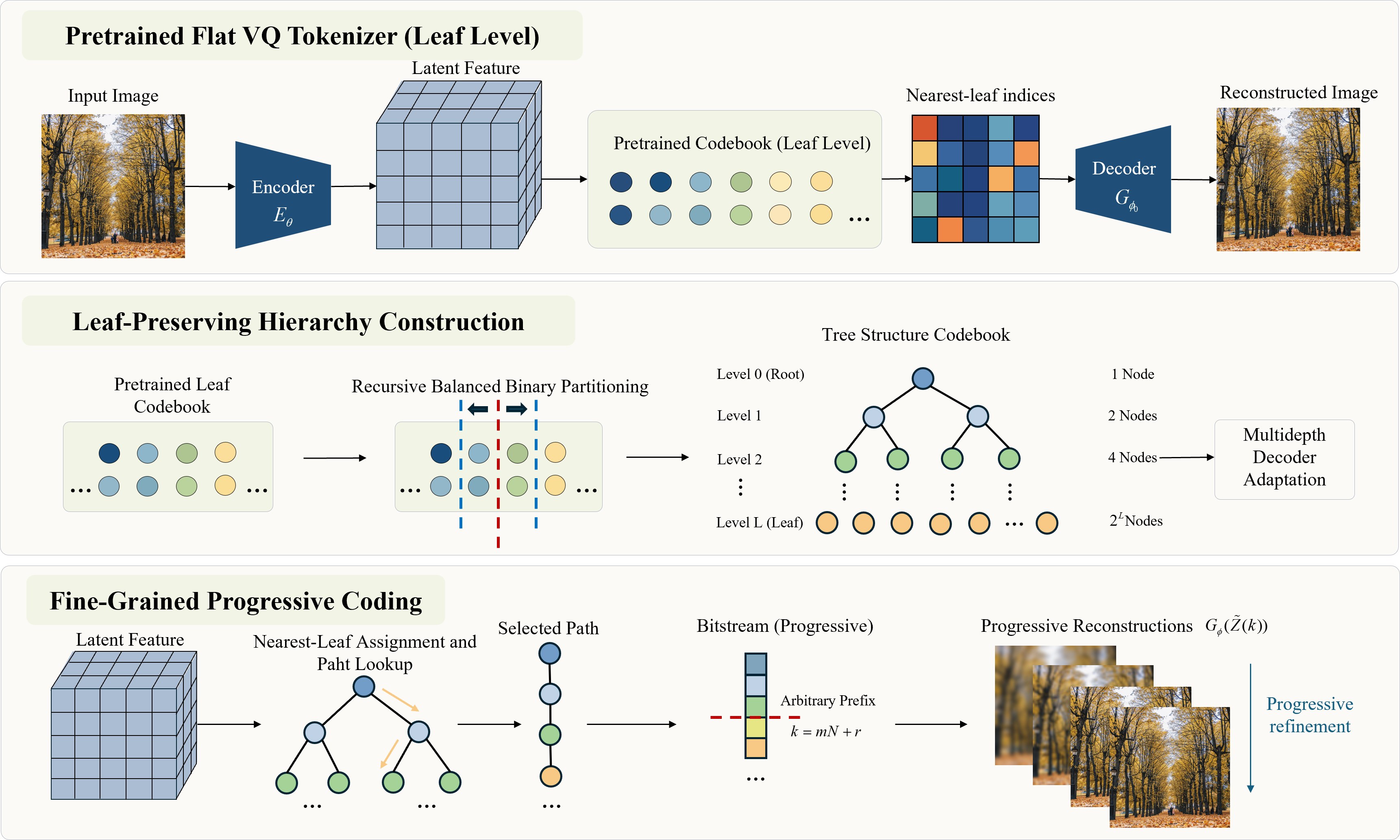}
	\caption{\method~overview. A pretrained flat codebook is reorganized into a balanced binary hierarchy while preserving all leaf vectors and encoder assignments. Internal nodes define intermediate latent states, and the decoder is adapted on the same complete-depth and mixed-depth prefixes encountered at inference.}
	\label{fig:overview}
\end{figure}

\subsection{Leaf-preserving hierarchy}
We recursively partition the $K$ leaves into equal-cardinality subsets until each subset contains one codeword. For each subset, two-means provides a separation direction; when the assignment is unbalanced, we project codewords onto the line between the two cluster centers and split at the median. This produces a complete binary tree of depth $L$ while only permuting the original leaves.

For an internal node $u$ with children $u_0$ and $u_1$, the default representative is
\begin{equation}
	c_u=\frac{1}{2}\left(c_{u_0}+c_{u_1}\right).
	\label{eq:parent}
\end{equation}
Because each subtree contains the same number of leaves, this recursively equals the unweighted mean of all descendant leaves. For conditions A--C in the hierarchy ablation, this rule initializes all internal representatives; only the leaf-to-path assignment changes. Conditions D and E keep the geometric topology, leaf assignment, and token order fixed, and optimize the internal-node representations while passing them through the same decoder. D adds distortion supervision to internal nodes; E continues from D and adds perceptual reconstruction supervision. The encoder, decoder, and leaf vectors are fixed during this internal-node optimization. Afterward, the optimized internal representations are frozen before the decoder-adaptation stage.

Let $a_d(j_i)$ denote the ancestor at depth $d$ on the path to assigned leaf
$j_i$, with $d=0$ at the root and $d=L$ at the leaf. At complete depth $d$,
token $i$ is represented by $c_{a_d(j_i)}$, and the shared decoder maps the
resulting token sequence to a reconstruction.

\subsection{Fine-grained prefix representation}
\label{sec:prefix}
Each selected leaf is represented by $L$ binary branch decisions
$b_i=(b_i^{(1)},\ldots,b_i^{(L)})$. We serialize these decisions in
depth-major order: transmit the $N$ first-level decisions in token order, then
the $N$ second-level decisions, and continue in this way through level $L$.
Thus each new branch decision refines exactly one token by one tree level.

\begin{figure}[htbp]
	\centering
	\includegraphics[width=\linewidth]{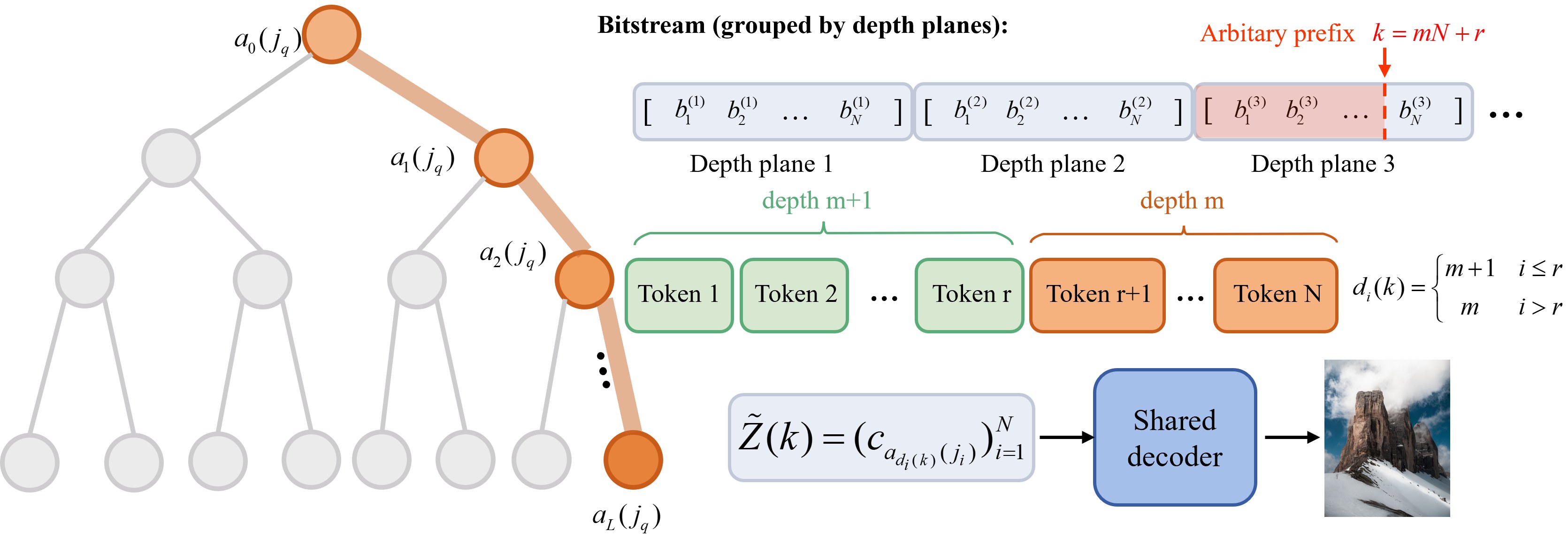}
	\caption{Mixed-depth latent representation induced by an arbitrary payload prefix.}
	\label{fig:mixedprefix}
\end{figure}

For a logical branch-prefix of length $k$, let $m=\lfloor k/N\rfloor$ be
the number of complete depth planes and $r=k-mN$ the number of received
branches in the next plane. As illustrated in Figure~\ref{fig:mixedprefix}, truncation inside a depth plane produces a mixed-depth latent state. The first $r$ tokens in the fixed serialization order are refined to depth $m+1$ and the
remaining $N-r$ tokens stay at depth $m$. The decoder represents each token by the ancestor
at its current depth on its fixed assigned-leaf path; at $k=LN$, every token
has reached its leaf. We denote the resulting mixed-depth token sequence by
$\widetilde Z(k)$.

Given the shared tree and image header, every logical branch-prefix
$k\in\{0,\ldots,LN\}$ uniquely determines a valid tree node for each token.
Depth-major serialization ensures that a token receives each branch only
after its preceding path is known; the prefix length fixes its current depth,
and the received decisions fix its node. Thus, for $k<LN$, appending one bit
refines exactly one token to one of its children. The full proof and
payload-state accounting are given in Appendix~\ref{app:proof}.

The guarantee above concerns the logical branch stream. With raw branch coding, every branch position is also a valid physical truncation point. If an entropy coder is used, the set of byte- or bit-level truncation points additionally depends on the coder and container format.

\subsection{Decoder adaptation for mixed-depth prefixes}
\label{sec:adapt}
A decoder trained only on leaf embeddings is not optimized for newly introduced ancestor vectors. After any internal-node optimization, we freeze the analysis transform, the projections into the quantization space, the tree topology, and all leaf and internal prototypes. We then optimize only the synthesis-side transform $G_\phi$, including its projections from quantized tokens back to decoder channels. In the default decoder-adaptation schedule, a discriminator is active only during the second (perceptual) stage.

For each training image, we sample a branch-prefix length $k\sim p(k)$, including positions inside a depth plane, construct $\widetilde Z(k)$, and retain full-depth supervision. We use the objective
\begin{equation}
\begin{aligned}
	\mathcal L_{\mathrm{dec}}
	=&\;\lambda_L\,\ell_L\!\left(x,G_\phi(\widetilde Z(LN))\right)
	+\mathbb E_{k\sim p(k)}\Big[
	\lambda_{\mathrm{mse}}\,\mathrm{MSE}(x,\hat x_k) \\
	&\hspace{24mm}
	+\lambda_{\mathrm{perc}}\,\mathrm{LPIPS}(x,\hat x_k)
	+\lambda_{\mathrm{adv}}(t)\,\ell_{\mathrm{adv}}(\hat x_k)
	\Big],
\end{aligned}
\label{eq:loss}
\end{equation}
where $\hat x_k=G_\phi(\widetilde Z(k))$, $\ell_L(x,\hat x)=\mathrm{MSE}(x,\hat x)+\mathrm{LPIPS}(x,\hat x)$, and $\ell_{\mathrm{adv}}(\hat x)=-D_\psi(\hat x)$ is the hinge generator loss for discriminator $D_\psi$. For decoder adaptation we set $\lambda_L=\lambda_{\mathrm{mse}}=\lambda_{\mathrm{perc}}=1$; $\lambda_{\mathrm{adv}}(t)$ is zero before update 5,000 and $0.02$ thereafter in the perceptual stage. For the internal-node ablation, let $\eta=\{c_u:u\text{ is internal}\}$
collect the $K-1$ internal vectors. Denote the fixed-decoder reconstruction by
$\hat x_{k,\eta}=G_{\phi_0}(\widetilde Z_\eta(k))$, where
$\widetilde Z_\eta(k)$ uses the internal prototypes in $\eta$. D initializes
$\eta$ with the recursive child means in
\cref{eq:parent} and optimizes sampled-prefix MSE. E initializes from D and
adds sampled-prefix LPIPS and the adversarial generator term, alternating
updates to $\eta$ and the discriminator as detailed in
Appendix~\ref{app:decoder-training}. In both conditions, topology, leaf
assignment, token order, encoder, leaf vectors, and decoder $G_{\phi_0}$ are
fixed. The full-depth term has no gradient with respect to $\eta$ because the
endpoint uses only frozen leaves. After this ablation, all tree vectors are
frozen before decoder adaptation.

The default $p(k)$ is defined on integer lengths $k\in\{0,\ldots,LN\}$ as the
rate-balanced discrete distribution below. It divides
the nonzero prefix lengths into six bins of equal width in $\log_2(k+N)$,
chooses a bin uniformly, then chooses an integer prefix uniformly within
that bin:
\begin{equation}
\begin{aligned}
	s_j &= \log_2(N+1)+\frac{j}{6}
	\left[\log_2((L+1)N)-\log_2(N+1)\right],
	\quad j=0,\ldots,6, \\
	\mathcal K_j &=
	\begin{cases}
	\{k\in\{1,\ldots,LN\}:s_{j-1}\leq\log_2(k+N)<s_j\},
		&j=1,\ldots,5,\\
	\{k\in\{1,\ldots,LN\}:s_5\leq\log_2(k+N)\leq s_6\},
		&j=6,
	\end{cases}\\
	p(k)&=\sum_{j=1}^{6}\frac{\mathbf{1}[k\in\mathcal K_j]}{6|\mathcal K_j|}.
\end{aligned}
\label{eq:prefix-pmf}
\end{equation}
Thus $p(0)=0$, so the root-only state is not sampled during training. The complete
sampler and batching schedule are detailed in Appendix~\ref{app:prefix-sampling}.
We compare fixed-depth, complete-plane, and rate-balanced prefix sampling in
\cref{sec:ablation} to test whether mixed-depth exposure is necessary.

\subsection{Rate and routing}
With uncoded binary branches, a logical prefix of $k$ decisions contributes $k/(HW)$ payload bpp. For a $256\times256$ image crop, one branch decision changes the raw payload rate by approximately $1.53\times10^{-5}$ bpp. In our default configuration, $L=12$ and $N=336$ for that reference crop, yielding $LN+1=4{,}033$ logical prefix states and a full raw branch payload of
\(12\times336/256^2=0.061523\) bpp. At full resolution, the token count and number of logical prefix states scale with $HW$, while the normalized payload rate remains the same.
Actual serialized rate is
\begin{equation}
	R_{\mathrm{actual}}(k)=
	\frac{|H_{\mathrm{img}}|+|C(B_{1:k})|+|P(k)|}{HW},
	\label{eq:actualrate}
\end{equation}
where $H_{\mathrm{img}}$ is the image header, $C$ is the chosen branch coder, and $P(k)$ denotes packing or container overhead. All lengths in this equation are measured in bits; byte counts are multiplied by eight. We report serialized bytes for RD evaluation.

Within each depth plane, the default token order is the fixed scale-and-raster order inherited from token concatenation. Our main results retain exhaustive nearest-leaf assignment, preserving the pretrained endpoint representation. Greedy tree routing reduces search from $O(KD)$ to $O(LD)$ comparisons per token, but may choose a different leaf, so we evaluate it separately as a quality--complexity trade-off.

\section{Experiments}
\label{sec:experiments}

\subsection{Setup and evaluation protocol}
\label{sec:setup}
\paragraph{Model and data.}
We train the base tokenizer on 300,000 Open Images crops
\citep{kuznetsova2020open}. Inspired by the architecture of RDVQ \citep{Jiang_2026_CVPR}, the autoencoder produces maps at strides
$16$, $32$, and $64$, projected to $D=32$; the shared $K=4096$ codebook gives
$L=12$ and 336 tokens for a $256\times256$ crop. We evaluate Kodak (24 images),
DIV2K, and CLIC2020 \citep{agustsson2017ntire,toderici2020workshop}. All results are
obtained using full-resolution inference.

\paragraph{Metrics and baselines.}
We report LPIPS~\citep{zhang2018unreasonable},
DISTS~\citep{ding2020image}, and CLIP-IQA~\citep{wang2023exploring}. For
locally evaluated models, DISTS and LPIPS are computed with the same pyiqa
implementation, and bpp is recomputed from locally generated bitstreams. The
perceptual plots include MS-ILLM, GLC, PerCo, RDEIC, DLF, OSCAR, OneDC,
RDVQ-adj, and our ProGIC reproduction (ProGIC*)
\citep{muckley2023improving,jia2024generative,careil2024towards,li2025rdeic,xue2025dlf,guo2025oscar,xue2025one,Jiang_2026_CVPR,Cao_2026_CVPR}.
All local inference and bitstream measurements use the same RTX 4090
environment. For BD-rate, we report positive rate savings over the shared
DISTS range and do not extrapolate~\citep{bjontegaard2001calculation}. The
plotted ProGIC* curve uses our local reproduction; BD-rate calculations use
the DISTS rate--quality points reported in the ProGIC paper, as detailed in
Appendix~\ref{app:baseline-provenance}.

\subsection{Prefix reconstruction and ablations}
\label{sec:ablation}
\begin{figure}[tbp]
	\centering
	\begin{subfigure}{0.45\textwidth}
		\centering
		\includegraphics[width=\linewidth]{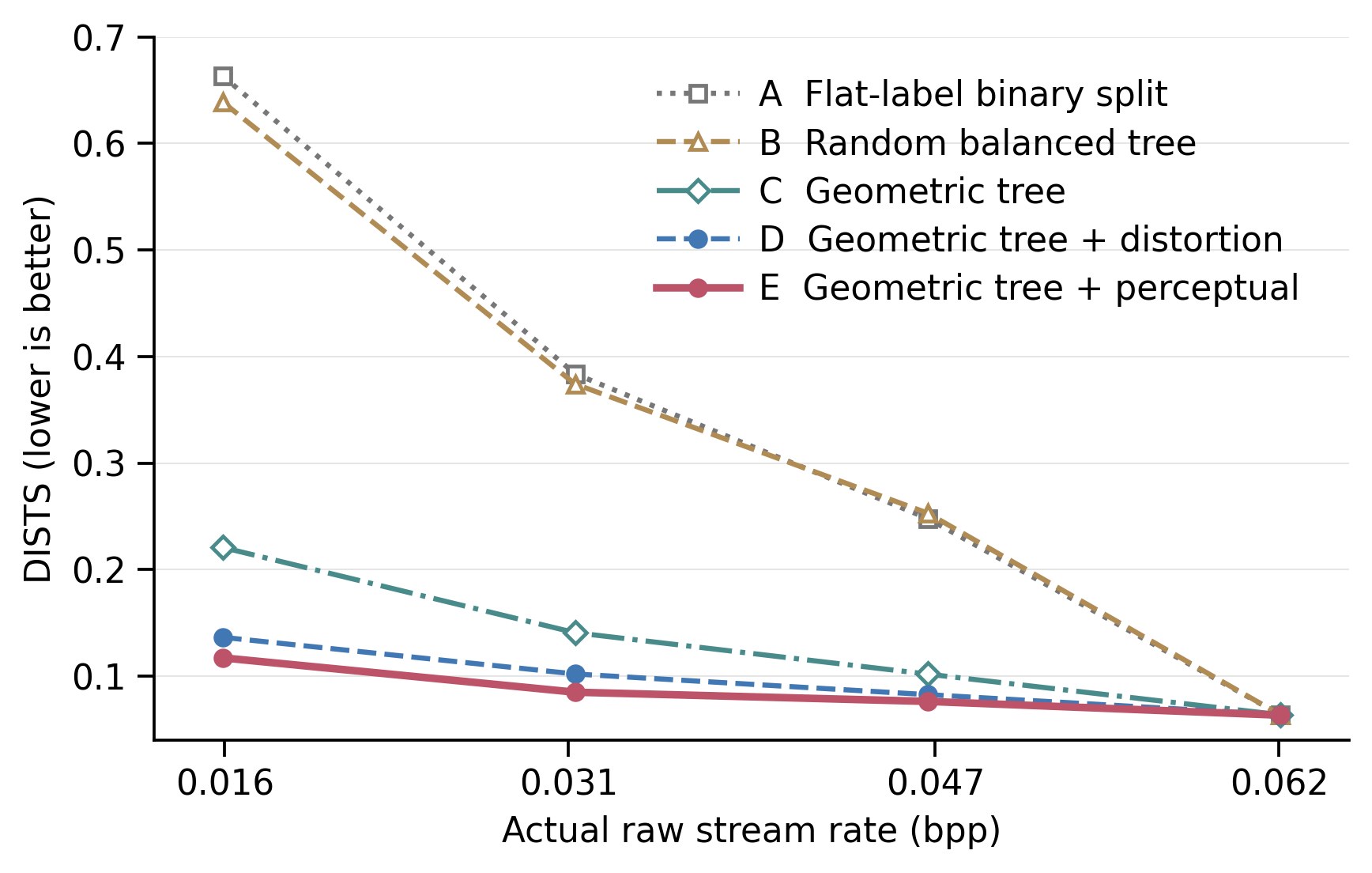}
		\caption{DISTS--rate curves.}
		\label{fig:ablation-geometry}
	\end{subfigure}
	\hfill
	\begin{subfigure}{0.45\textwidth}
		\centering
		\includegraphics[width=\linewidth]{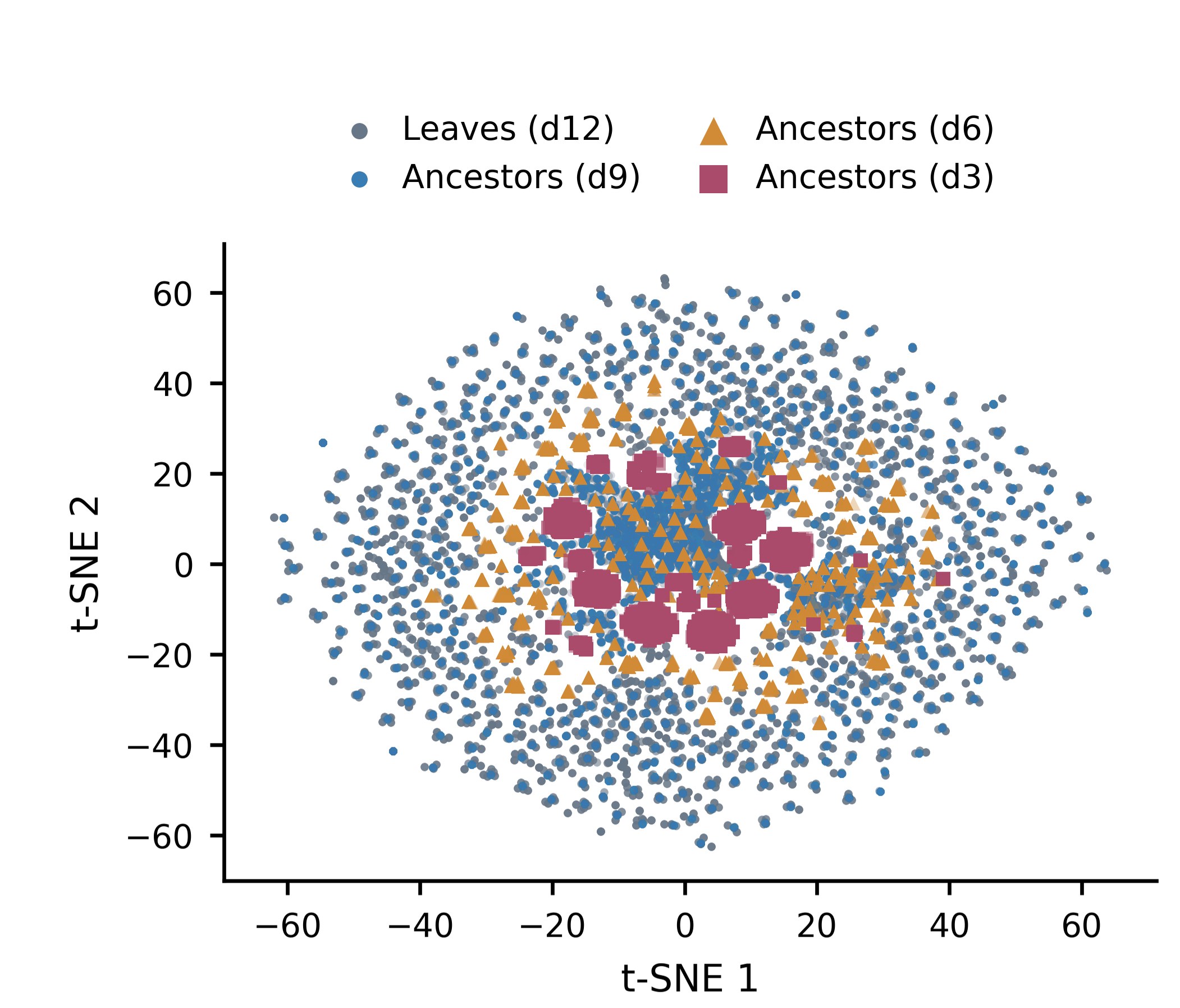}
		\caption{t-SNE of token vectors.}
		\label{fig:ablation-shift}
	\end{subfigure}
	\caption{Controlled path-assignment and internal-node-supervision ablations.
	(a) DISTS versus raw branch rate for five conditions built from the same
	pretrained tokenizer and frozen leaves.  (b) t-SNE visualization of leaf and
	ancestor representations at the indicated depths.}
	\label{fig:component-ablation}
\end{figure}

Figure~\ref{fig:component-ablation} separates path assignment from
internal-node reconstruction. Conditions A--C use the same pretrained
encoder, leaf vectors, decoder, node-representative rule, and token order;
they vary only the leaf-to-path assignment: the original flat-label order, a
random balanced permutation, or the recursively clustered geometric order.
C, D, and E use the same geometric topology, leaf assignment, branch
serialization, and decoder. C uses the recursive child-mean internal
vectors. D optimizes only the internal-node vectors with sampled-prefix MSE
supervision; E initializes from D and adds LPIPS and adversarial
reconstruction supervision. In D/E the decoder is held fixed while internal
vectors are optimized; the encoder, leaf vectors, tree topology, and token
order also remain fixed. The plotted C-to-D-to-E progression shows the
effect of progressively adding reconstruction supervision to internal nodes
under the stated optimization stages.
The t-SNE panel shows that ancestor states occupy a distribution distinct
from the leaf states used by the original decoder.

\begin{figure}[tbp]
	\centering
	\includegraphics[width=\linewidth]{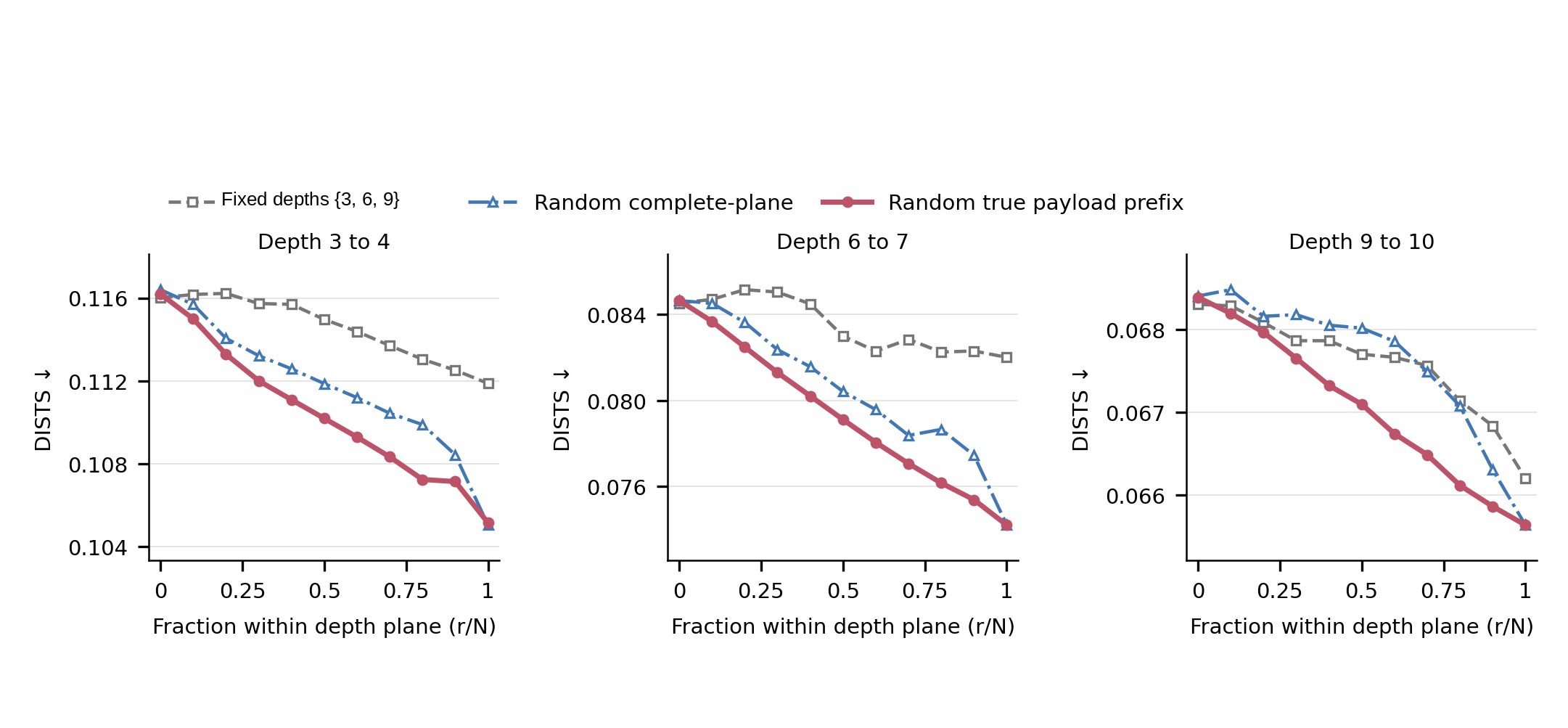}
	\caption{Prefix-sampling ablation on Kodak. DISTS is shown during the
	$3\!\to\!4$, $6\!\to\!7$, and $9\!\to\!10$ transitions; $r/N$ is the
	fraction of tokens advanced to the next depth. Red denotes rate-balanced
	random payload-prefix sampling. Blue samples only complete depths
	$d\in\{0,\ldots,L\}$, whereas gray uses only the fixed diagnostic depths
	$d\in\{3,6,9\}$. Only the red condition samples within-plane mixed-depth
	states.}
	\label{fig:payload}
\end{figure}

The decoder must also see mixed-depth states. On Kodak,
Fig.~\ref{fig:payload} shows that rate-balanced random payload-prefix sampling
gives the lowest DISTS at intermediate $r/N$ in all three depth transitions.
The complete-plane baselines lack this intermediate supervision, supporting
training on the prefixes encountered during transmission. Appendix
\ref{app:ablations} reports measured within-plane token-ordering results.

\subsection{Rate--quality and qualitative comparisons}
\label{sec:perceptual-rd}
\begin{figure}[tbp]
	\centering
	\includegraphics[width=\linewidth]{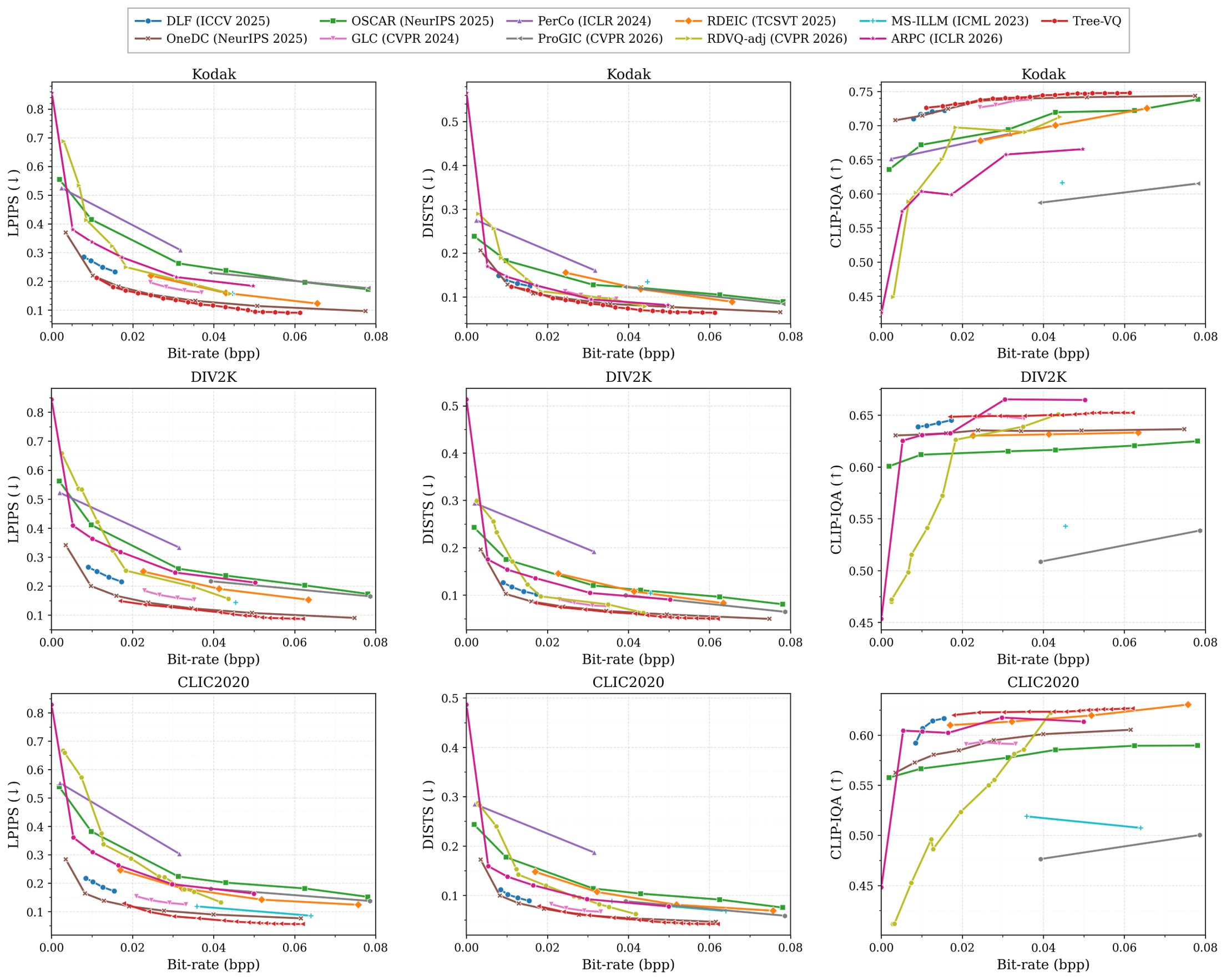}
\caption{Perceptual rate--quality comparison over the compatible low-rate
range. Rows are Kodak, DIV2K, and CLIC2020; columns are LPIPS (lower),
DISTS (lower), and CLIP-IQA (higher). The red Tree-VQ curve contains
evaluated prefix states; the ProGIC* curve is from our local reproduction.}
	\label{fig:perception}
\end{figure}

Figure~\ref{fig:perception} reports LPIPS, DISTS, and CLIP-IQA on
Kodak, DIV2K, and CLIC2020 over the overlapping low-rate range. Tree-VQ
achieves the lowest LPIPS/DISTS over most of the plotted range and among the
highest CLIP-IQA scores. The ProGIC* curve represents our local reproduction.

\begin{figure}[tbp]
\centering
\includegraphics[width=\linewidth]{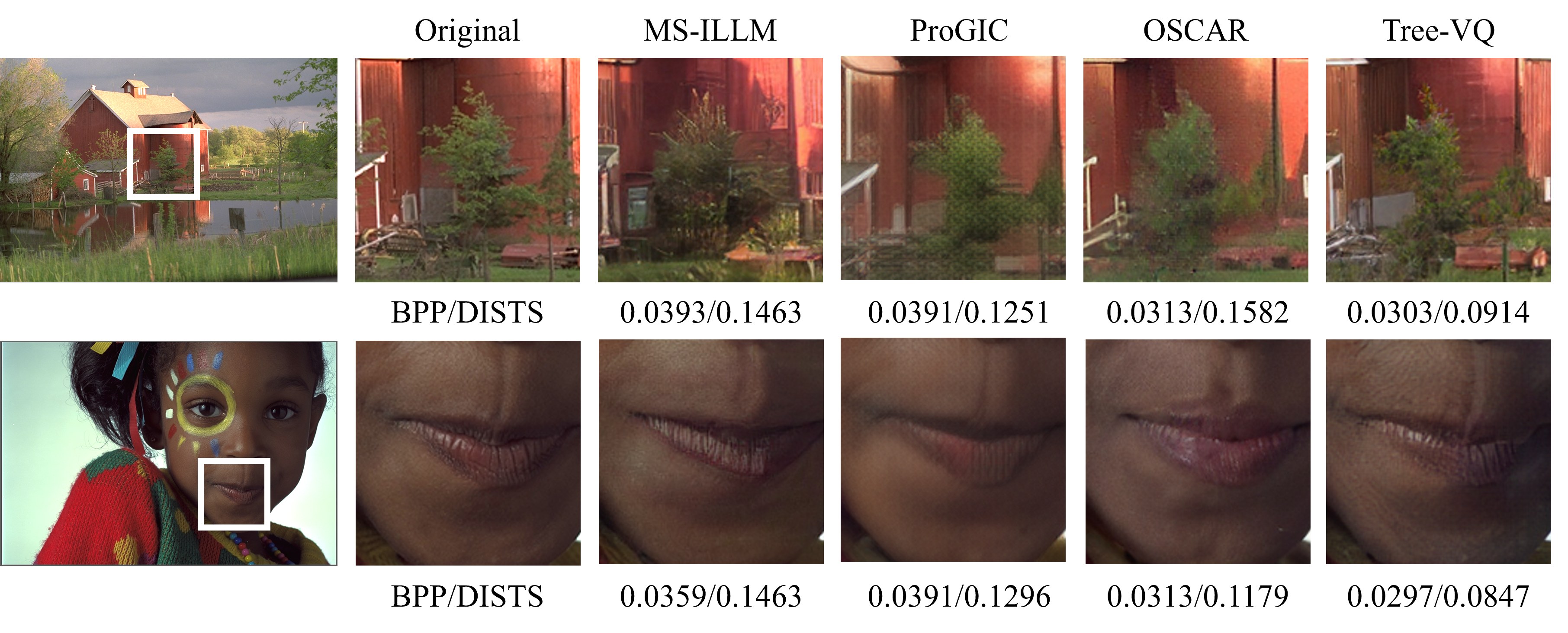}
\caption{Qualitative comparisons on two images. In each row, the left panel
shows the source image with the enlarged region marked by a white box; the
remaining panels show the original crop followed by reconstructions from
MS-ILLM, ProGIC, OSCAR, and Tree-VQ. Each method panel is annotated with its
per-image bpp and DISTS value (lower is better).}
\label{fig:tree_recon1}
\end{figure}

The qualitative preview in Fig.~\ref{fig:tree_recon2} shows the original and
Tree-VQ reconstruction alongside crops from OneDC, MS-ILLM, RDEIC, and OSCAR.
Figure~\ref{fig:tree_recon1} provides two further comparisons, with per-image
bpp and DISTS reported for each method. Additional Tree-VQ reconstructions are
in Appendix~\ref{app:qualitative}.

\subsection{Coding and efficiency}
\label{sec:entropy-analysis}
\begin{figure}[H]
\centering
\begin{subfigure}[t]{0.37\linewidth}
\centering
\includegraphics[width=\linewidth]{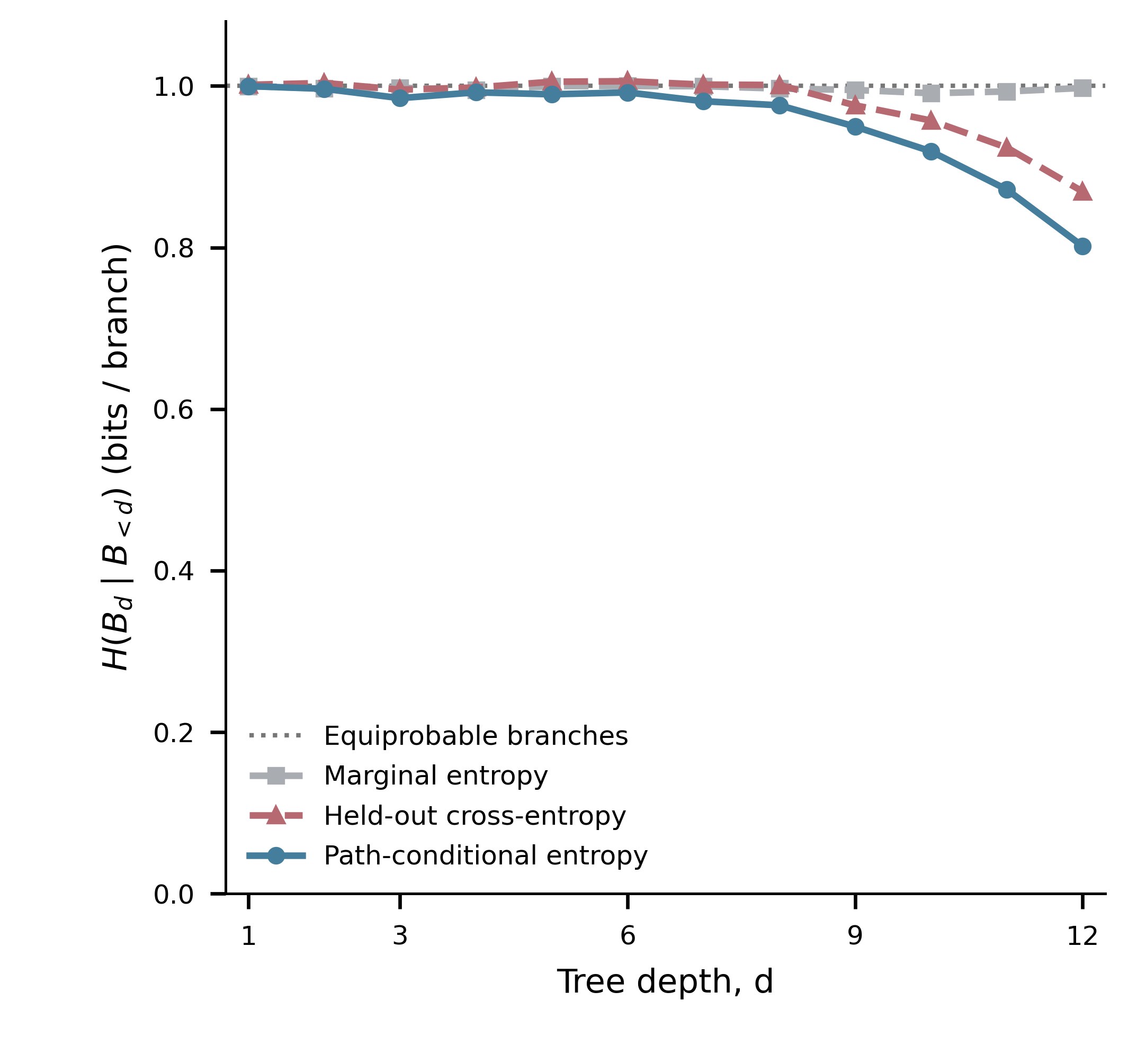}
\caption{Depth-wise branch entropy.}
\label{fig:branch-entropy}
\end{subfigure}
\hfill
\begin{subfigure}[t]{0.515\linewidth}
\centering
\includegraphics[width=\linewidth]{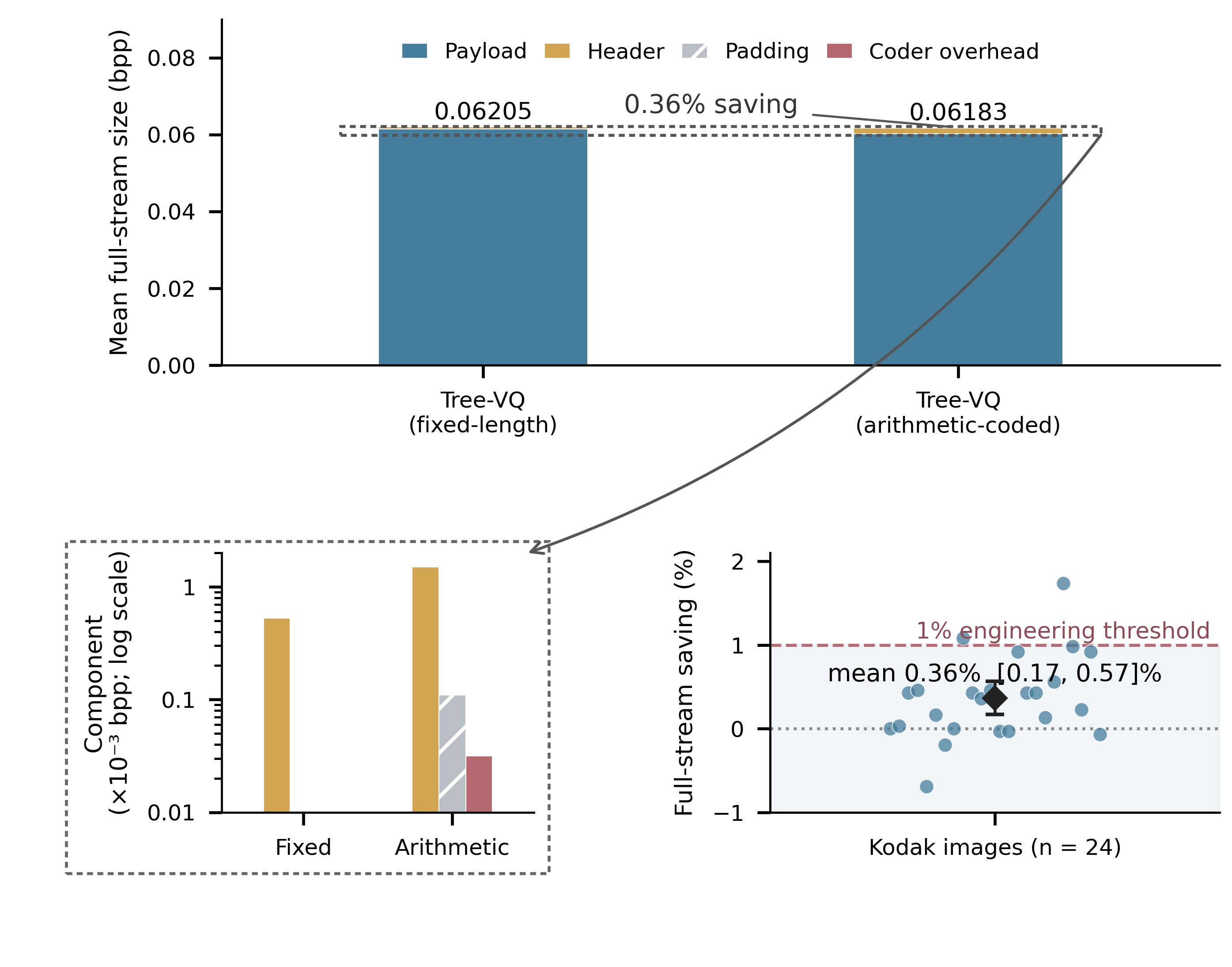}
\caption{Serialized-rate breakdown on Kodak.}
\label{fig:stream-coding}
\end{subfigure}
\caption{Depth-wise path-conditional branch entropy and serialized-rate
comparison on Kodak. The arithmetic-coded rate includes image-header and
container overhead.}
\label{fig:entropy-coding}
\end{figure}

At depth 12, path-conditional branch entropy is $0.8$ bit. On Kodak,
arithmetic coding reduces the serialized full-stream rate from $0.06205$ to
$0.06183$ bpp, a $0.36\%$ reduction after image-header and container overhead.
We therefore use fixed-length branches in the reported rate--distortion
results. Figure~\ref{fig:entropy-coding} shows the depth-wise entropy and
serialized-rate breakdown; routing and runtime details are in
Appendix~\ref{app:rate-runtime}.

\FloatBarrier
\section{Conclusion}
\label{sec:conclusion}

We presented Tree-VQ, a post-hoc progressive coding framework for a fixed pretrained VQ tokenizer. It preserves the encoder assignments and learned leaf vectors, organizes the codebook into a balanced binary hierarchy, and assigns explicit representations to internal nodes. Depth-major transmission makes each payload prefix a valid mixed-depth latent state. A shared adapted decoder maps these states to reconstructions while preserving the pretrained tokenizer's full-depth latent sequence.  

On Kodak, DIV2K, and CLIC2020, Tree-VQ achieves DISTS BD-rate savings of 52.1\%, 49.0\%, and 53.4\%, respectively, relative to the published ProGIC points. The fixed image header and received payload length let the decoder recover a stopped prefix of the same stream, so useful reconstructions can arrive as bits are transmitted without re-encoding for each target rate. These results show that bit-level progressive control can coexist with strong perceptual rate–distortion performance at ultra-low rates. More broadly, Tree-VQ adds fine-grained transmission to a pretrained neural tokenizer through post-hoc hierarchy construction and shared-decoder adaptation.

\subsection*{AI use statement}
We used generative AI tools to polish manuscript prose, including suggestions
for clearer phrasing, grammar, and concision. We also used them to support
literature discovery by suggesting search terms and candidate related papers.
The authors assessed the relevance of the suggested literature and checked
citations against the original publications. All AI-suggested text was reviewed
and edited by the authors, who take responsibility for the final manuscript,
including its claims and references.

\subsection*{Ethics statement}
This work does not involve human participants or collect new personal data. We
use existing image datasets and introduce no new dataset.

\subsection*{Reproducibility statement}
Section~\ref{sec:method} describes the Tree-VQ representation and decoder
adaptation, while Section~\ref{sec:setup} specifies the evaluation datasets,
metrics, and baseline sources. The appendix provides the tokenizer architecture
and training details (Appendices~\ref{app:architecture} and
\ref{app:decoder-training}), hierarchy construction and prefix sampling
(Appendices~\ref{app:architecture} and \ref{app:prefix-sampling}), evaluation
preprocessing and rate accounting (Appendix~\ref{app:protocol}), the
prefix-decodability proof (Appendix~\ref{app:proof}), and additional ablations
and runtime measurements (Appendices~\ref{app:ablations} and
\ref{app:rate-runtime}).

\bibliography{iclr2027_conference}
\bibliographystyle{iclr2027_conference}

\appendix

\section{Architecture and Training Details}
\subsection{Base tokenizer architecture}
\label{app:architecture}
Based on RDVQ~\citep{Jiang_2026_CVPR}, our tokenizer uses a multi-scale
convolutional autoencoder.  The analysis and synthesis transforms have seven
feature resolutions with channel widths
$(128,128,256,256,512,512,512)$.  The analysis transform produces latent maps
at strides $16$, $32$, and $64$, with spatial sizes $16^2$, $8^2$, and $4^2$
for a $256\times256$ training crop.  Each map has 256 channels before being projected
by an independent learned $1\times1$ convolution to the common token
dimension $D=32$.  The three scales share a binary tree with 4096 leaves and
depth $L=12$.  Thus,
\begin{equation}
	N=\frac{256^2}{16^2}+\frac{256^2}{32^2}+\frac{256^2}{64^2}
	=256+64+16=336
\end{equation}
tokens are produced per $256\times256$ training crop. At inference, the same
transforms operate on the complete image at its native resolution.

Residual blocks use GroupNorm with 32 groups, SiLU activations, and two
$3\times3$ convolutions.  Down-sampling uses stride-2 $3\times3$
convolutions, whereas up-sampling uses nearest-neighbour $2\times$
interpolation followed by a $3\times3$ convolution.  Global spatial
self-attention is applied at the three lowest-resolution feature maps
($16^2$, $8^2$, and $4^2$) in both transforms and in the corresponding
scale-fusion blocks.  After quantization, each scale is projected back to 256
channels and added to the synthesis transform at its matching resolution.

\begin{table}[ht]
	\centering
	\caption{Base tokenizer architecture.  ``RB'' denotes a two-convolution
		residual block and ``Attn'' denotes one global spatial self-attention
		block.}
	\label{tab:arch}
	\scriptsize
	\setlength{\tabcolsep}{2pt}
	\begin{tabular}{@{}p{0.18\linewidth}p{0.15\linewidth}p{0.13\linewidth}p{0.28\linewidth}p{0.16\linewidth}@{}}
		\toprule
		Stage & Resolution & Channels & Main operations & Output \\
		\midrule
		Stem & $256^2$ & $128$ & $3\times3$ convolution & -- \\
		Analysis levels & $256^2$--$32^2$ & $128/256$ & $2$ RB; stride-2 between levels & $16^2$ feature \\
		Low-resolution analysis & $16^2,8^2,4^2$ & $512$ & $2$ RB $+2$ Attn per level & three scale features \\
		Scale heads & $16^2/8^2/4^2$ & $512\!\rightarrow\!256$ & $2$ RB $+$ Attn $+$ GN $+$ $3\times3$ conv. & latent maps \\
		Token projection & $16^2/8^2/4^2$ & $256\!\rightarrow\!32$ & $1\times1$ conv. per scale & quantized tokens \\
		\midrule
		Synthesis scale inputs & $4^2/8^2/16^2$ & $32\!\rightarrow\!256\!\rightarrow\!512$ & $1\times1+3\times3$ conv.; $2$ RB $+$ Attn & fused features \\
		Synthesis levels & $4^2$--$256^2$ & $512/256/128$ & $3$ RB per level; Attn at $4^2$--$16^2$ & RGB feature \\
		Output & $256^2$ & $128\!\rightarrow\!3$ & GN $+$ SiLU $+$ $3\times3$ conv. & RGB image \\
		\bottomrule
	\end{tabular}
\end{table}

The analysis transform contains 80,462,848 parameters and the synthesis
transform contains 105,159,555.  The six scale projections together contain
50,016 parameters.  The tree stores all $2\times4096-1=8191$ node prototypes
in 32 dimensions: 4096 leaves contribute 131,072 parameters and 4095 internal
nodes contribute 131,040 parameters.  Although the internal prototypes are
initialized from the leaf hierarchy, they are represented explicitly in the
model.  The complete tokenizer therefore contains
185,934,531 parameters (185.934531 million), excluding training-only modules
and non-parameter buffers.

\subsection{Hierarchy construction}
Within every current leaf subset $S$, two-means is initialized deterministically: the first center is the codeword farthest from the subset mean and the second is the codeword farthest from the first center. We run 8 Lloyd assignment/update iterations or stop earlier when assignments do not change. If the resulting child sets have unequal cardinality, all vectors are projected onto the line joining the two temporary centroids and split at the median projection value. Ties are broken by the original codebook index, making the topology deterministic for a fixed leaf vocabulary.

Conditions A--C in \cref{fig:component-ablation} differ only in the
assignment of the same leaf vectors to binary paths. For A, leaf $j$ follows
the length-$L$ binary expansion of $j-1$, most-significant bit first;
equivalently, $b_j^{(d)}=\lfloor(j-1)/2^{L-d}\rfloor\bmod 2$ for
$d=1,\ldots,L$. For B, we draw a uniform permutation $\pi$ of the $K$ leaves
once; leaf $c_{\pi(t)}$ follows the length-$L$ binary expansion of $t-1$,
most-significant bit first, for $t=1,\ldots,K$. A branch bit of 0 selects the
left child and a bit of 1 selects the right child. C uses the recursive
geometric partition described above.
All three use the same recursive child-mean rule in \cref{eq:parent} and the
same fixed scale-and-raster token order. D and E use C's exact topology,
leaf-to-path assignment, and token order. During both stages, the pretrained
decoder is held fixed and the generator-side trainable parameters are only
the $K-1$ internal-node vectors; the encoder, encoder-side projections, all
leaf vectors, and topology are frozen. D starts from C's child-mean vectors
and uses sampled-prefix MSE supervision. E starts from D's internal vectors
and adds sampled-prefix LPIPS and a weak hinge-GAN term; its discriminator is
the only additional trainable module. After this ablation, the resulting
node vectors are frozen before the common decoder-adaptation procedure.

All five conditions use the same 300,000-crop Open Images training set,
random $256\times256$ crops, and Kodak evaluation protocol. A--C use their
specified fixed internal vectors. For D/E, the rate-balanced prefix sampler
in Appendix~\ref{app:prefix-sampling} selects supervision states and the
pretrained decoder is held fixed while only internal-node vectors are
optimized. D uses 180,000 updates, global batch size 16, and Adam with learning
rate $1\times10^{-5}$. E starts from D's internal vectors and continues for
122,000 updates with the counter restarted, global batch size 32, and Adam
with learning rate $5\times10^{-5}$ for internal vectors and
$2\times10^{-5}$ for the discriminator. The discriminator is reset at the
start of E; its weight is zero for the first 5,000 updates and $0.02$
thereafter. Both stages use $(\beta_1,\beta_2)=(0.9,0.95)$, with no weight
decay or learning-rate schedule. Figure~\ref{fig:component-ablation} DISTS
curves decode all representations
with the same frozen pretrained decoder, before subsequent decoder
adaptation. C-to-D adds distortion supervision to internal vectors;
D-to-E adds perceptual/adversarial supervision and uses the second-stage
optimization schedule.

\subsection{Decoder adaptation}
\label{app:decoder-training}
The analysis transform, the projections into the quantization space, and all
tree prototypes are kept fixed after hierarchy conversion and internal-node
optimization. The synthesis-side transform, including the projections from
quantized tokens back to the decoder channels, is optimized. An adversarial
discriminator is introduced only in the second stage.

During the hierarchy ablation, D and E instead hold the pretrained decoder
fixed and optimize only the $K-1$ internal vectors of the geometric tree.
The topology, leaf-to-path assignment, token order, encoder, encoder-side
projections, and $K$ leaf vectors are frozen. D minimizes the sampled-prefix
MSE term of \cref{eq:loss}. E is initialized from D and alternates generator
updates on $\eta$ using the sampled-prefix LPIPS and adversarial terms with
discriminator updates on $\psi$. The full-depth loss in
\cref{eq:loss} is retained, but has zero gradient with respect to internal
vectors because full depth uses only frozen leaves. The resulting internal
vectors are then frozen before the shared decoder-adaptation procedure.

For the main model, after hierarchy construction (and any separately
evaluated internal-node ablation), we use two successive decoder-adaptation
stages on $256\times256$ training crops. Each stage is allocated an upper budget of
600,000 updates. The reported checkpoints are at 18,000 updates in the first
stage and 122,000 updates in the second stage (with the second-stage counter
restarted at zero). The
first stage uses a global batch size of 16 and Adam with learning rate
$1\times10^{-5}$ and
$(\beta_1,\beta_2)=(0.9,0.95)$.  The second stage uses a global batch size of
32; the synthesis transform and discriminator use learning rates
$5\times10^{-5}$ and $2\times10^{-5}$, respectively, with the same Adam
coefficients.  No explicit weight decay, warm-up, or cosine decay is applied.
For the perceptual model, checkpoint selection uses a joint perceptual
criterion based on LPIPS and DISTS; we report the checkpoint with the lowest
combined score.

Both stages optimize the decoder objective in \cref{eq:loss}. The full-depth
term remains active throughout, while the prefix expectation samples arbitrary
nonzero logical lengths from $p(k)$, including positions inside a depth plane.
Thus the supervised prefix states are not restricted to the complete depths
$d\in\{3,6,9\}$; those depths are useful diagnostics but are not the only
training states. Terms involving the frozen analysis transform, quantization
projections, and tree prototypes are constant with respect to the optimized
decoder and are not shown here.

In the second stage, the generator objective remains \(\mathcal{L}_{\mathrm{dec}}\)
from \cref{eq:loss}, with
\begin{equation}
	\alpha_t=\begin{cases}
		0, & t<5{,}000,\\
		0.02, & t\ge 5{,}000.
	\end{cases}
\end{equation}
where \(\lambda_{\mathrm{adv}}=0\) in the first stage and
\(\lambda_{\mathrm{adv}}=\alpha_t\) in the second stage. The adversarial term is
therefore applied to the sampled prefix reconstructions in the same expectation
as Eq.~\eqref{eq:loss}. The discriminator is a five-convolution PatchGAN
operating on RGB crops and uses the hinge objectives
\begin{align}
	\mathcal{L}_{D}
	&=\frac{\alpha_t}{2}\left(
	\mathbb{E}_{x}[\max(0,1-D_\psi(x))]
		+\mathbb{E}_{k\sim p(k)}[\max(0,1+D_\psi(\hat{x}_{k}))]\right),\\
	\mathcal{L}_{G}^{\mathrm{adv}}
	&=-\mathbb{E}_{k\sim p(k)}[D_\psi(\hat{x}_{k})].
\end{align}
For E in the internal-node ablation, the discriminator uses the same hinge
loss with the fixed decoder and $\hat{x}_{k,\eta}$:
\begin{equation}
\mathcal{L}_{D,\mathrm{node}}(\psi;\eta)
=\frac{\alpha_t}{2}\left(
\mathbb{E}_{x}[\max(0,1-D_\psi(x))]
+\mathbb{E}_{k\sim p(k)}[\max(0,1+D_\psi(\hat{x}_{k,\eta}))]
\right).
\end{equation}
The generator update minimizes the sampled-prefix MSE, LPIPS, and adversarial generator terms over $\eta$
with $\psi$ fixed; the discriminator update uses the hinge loss with $\eta$
fixed.
The discriminator is reinitialized at the beginning of the second stage, and
the full-depth term is retained throughout to preserve the endpoint
reconstruction while the decoder learns to interpret internal tree nodes.

\subsection{Prefix sampling and batching}
\label{app:prefix-sampling}
The main model uses the distribution in \cref{eq:prefix-pmf}, which mildly
over-samples early prefixes compared with uniform sampling over
$\{0,\ldots,LN\}$ while covering the full nonzero range. Each sampled $k$ is
used in all prefix terms of \cref{eq:loss}, so positions inside a depth plane
produce the mixed-depth states used during deployment. In a mixed optimization
step, half of the images are evaluated at full depth and half at sampled
payload prefixes; every fourth step is an all-full-depth minibatch. This
schedule supplies both terms of \cref{eq:loss} without evaluating both
representations for every image and keeps endpoint gradients present
throughout training.

For the sampler ablation, the complete-plane baseline samples uniformly from
$k\in\{0,N,\ldots,LN\}$, the fixed-depth baseline samples uniformly from
$k\in\{3N,6N,9N\}$, and the payload-prefix baseline uses the rate-balanced
distribution above. All three runs share the same initialization, optimizer
budget, augmentations, and full-depth update schedule.

\section{Evaluation Protocol and Reproducibility}
\label{app:protocol}

\subsection{Full-resolution inference and reconstruction}
All results are obtained using full-resolution inference. Each source image is
color-decoded, orientation-corrected, and passed to the encoder at its native
spatial resolution. We do not resize, partition, or overlap-blend images at
evaluation time. If an image dimension is not compatible with the encoder's
stride, only the minimal boundary padding needed for that single full image is
applied and removed after decoding. Reconstructions and quality metrics are
computed on the complete original image. All locally evaluated bitrates are
recomputed from the produced bitstream bytes and divided by the original image
area $HW$. All local DISTS and LPIPS measurements use the same pyiqa
implementation.

\subsection{Fixed-stream truncation and full-image serialization}
Each image is encoded as one fixed-header stream. The header is written once
and records the original image geometry, any stride-alignment padding, and the
full payload length; these fields are identical for every prefix. To decode a
prefix, the receiver keeps the first $q$ payload bits of this stream. The
received bit length $q$ is obtained from the transport/container framing,
including the number of valid bits in a partially received final byte. No
field in the image header is rewritten when the stream stops. Padding bits in
the final byte, if present in a byte-oriented container, are excluded using
this received-length information. A byte-only file cut exposes only
byte-aligned values of $q$; finer bit-level cuts require the framing layer to
expose the valid-bit count.

Let $b_j$ denote branch position $j$ in the full-image depth-major stream,
where $j\in\{0,\ldots,LN-1\}$ and
$N=HW/16^2+HW/32^2+HW/64^2$ for the inference resolution. The image payload
serializes the branch positions in their fixed scale-and-raster token order:
\[
 b_0,b_1,\ldots,b_{LN-1}.
\]
Consequently, a prefix of $q=k$ bits gives the complete image token sequence
its first $k$ branch decisions, and every logical length
$k\in\{0,\ldots,LN\}$ is a valid full-image reconstruction state.

For a $512\times768$ Kodak image, full-resolution inference produces
$N=2016$ tokens and $LN=24192$ branch bits. The stream therefore exposes
24193 logical bit-level prefix lengths and 3025 byte-aligned payload lengths.
These counts exclude the fixed image header, which adds the same size to every
prefix. Images with other native resolutions use the corresponding value of
$N$; no cross-image tile interleaving is performed.

\subsection{Baseline provenance and comparison scope}
\label{app:baseline-provenance}
Table~\ref{tab:baseline-provenance} distinguishes locally evaluated checkpoints,
our ProGIC reproduction, and DISTS points taken from the ProGIC paper. Official
weights for ProGIC were not available when this paper was submitted, so the
locally reproduced model is a separate set of weights and is denoted
ProGIC*. It supplies the ProGIC curve in Fig.~\ref{fig:perception} and the
ProGIC* point in Fig.~\ref{fig:model-efficiency}. The headline DISTS BD-rate
comparisons instead use the DISTS rate--quality points reported in the original
ProGIC paper~\citep{Cao_2026_CVPR}.

\begin{table}[ht]
\centering
\caption{Sources used for the reported rate--quality results. ProGIC* denotes
our local reproduction and is distinct from the ProGIC data reported in its
paper.}
\label{tab:baseline-provenance}
\small
\begin{tabularx}{\linewidth}{p{0.21\linewidth}p{0.36\linewidth}X}
\toprule
Method/result & Checkpoint or data source & Use in this paper \\
\midrule
Tree-VQ & Our trained checkpoint; evaluated locally on RTX 4090 & Tree-VQ rate--quality curves and locally measured points for BD-rate comparisons \\
Other baselines & Official publicly released checkpoints; evaluated locally on RTX 4090 & Rate--quality plots and local comparisons; DISTS/LPIPS use pyiqa and bpp is recomputed from bitstreams \\
ProGIC* & Our local reproduction; original official weights unavailable at submission & ProGIC curve in Fig.~\ref{fig:perception} and ProGIC* point in Fig.~\ref{fig:model-efficiency} \\
ProGIC paper points & DISTS rate--quality data reported in the original paper~\citep{Cao_2026_CVPR} & DISTS BD-rate reference only; not re-evaluated locally \\
\bottomrule
\end{tabularx}
\end{table}

The published-ProGIC BD-rate savings are summarized in
Table~\ref{tab:published-progic-bdrate}. These are cross-paper comparisons:
Tree-VQ points use our local bitstream and metric pipeline, while the ProGIC
reference points retain the source paper's reported protocol. The locally
plotted ProGIC* curve is not the reference used for these three numbers.

\begin{table}[ht]
\centering
\caption{DISTS BD-rate savings of Tree-VQ relative to rate--quality points
reported in the ProGIC paper. Positive values indicate bitrate savings.}
\label{tab:published-progic-bdrate}
\small
\begin{tabular}{lrr}
\toprule
Dataset & Reference points & DISTS BD-rate saving \\
\midrule
Kodak & ProGIC paper~\citep{Cao_2026_CVPR} & $52.1\%$ \\
DIV2K & ProGIC paper~\citep{Cao_2026_CVPR} & $49.0\%$ \\
CLIC2020 & ProGIC paper~\citep{Cao_2026_CVPR} & $53.4\%$ \\
\bottomrule
\end{tabular}
\end{table}

\section{Proof of Payload-Prefix Decodability}
\label{app:proof}
Let $N$ be the number of tokens and $L$ the tree depth. The payload uses the depth-major serialization described in Sec.~\ref{sec:prefix}. For a received length $k$, write $m=\lfloor k/N\rfloor$ and $r=k-mN$, with $0\le r<N$. The first $r$ tokens in the next plane are at depth $m+1$ and the rest are at depth $m$.

At $k=0$, every token is at the root and therefore has a unique valid state.
Assume every token has a unique current node after $k$ bits, for some $k<LN$.
The $(k+1)$-st bit belongs to token $r+1$ in depth plane $m+1$. When
$r=N-1$, this is token $N$ and completes that plane; the following bit belongs
to token 1 of the next plane. Because all earlier depth planes for the selected
token have already been received, its current node is the unique depth-$m$
ancestor on its leaf path. The new branch bit selects exactly one of that
node's two children and leaves all other token states unchanged. Therefore, by
induction, every $k\in\{0,\ldots,LN\}$ determines exactly one current node
for every token. The fixed scale/spatial layout then converts these nodes into
a complete latent tensor for $G_\phi$.

\paragraph{Corollary: number of valid payload states.}
For a full-resolution image with $N$ tokens, the $LN$ branch decisions induce
$LN+1$ logical prefix states, including the root-only state. Physical bit-level
and byte-aligned payload counts for the complete image stream are summarized
in Appendix~\ref{app:protocol}.

\section{Additional Ablations}
\label{app:ablations}

\subsection{Within-plane token ordering}
Table~\ref{tab:ordering} compares measured DISTS at matched payload fractions
for a representative middle-depth transition. Image-adaptive ordering gives
the lowest DISTS, while a static importance order retains much of the gain
without image-specific side information.

\begin{table}[ht]
\centering
\caption{Measured within-plane ordering results at matched payload fractions
for a representative middle-depth transition. DISTS is lower-is-better; side
information is included in the rate accounting.}
\label{tab:ordering}
\small
\begin{tabular}{lcccc}
\toprule
Ordering & 25\% plane & 50\% plane & 75\% plane & Side info / image \\
\midrule
Scale + raster & 0.096 & 0.091 & 0.086 & 0 bits \\
Deterministic interleave & 0.094 & 0.089 & 0.085 & 0 bits \\
Static importance order & 0.092 & 0.087 & 0.083 & 0 bits \\
Image-adaptive importance & 0.090 & 0.084 & 0.081 & 84 bits \\
\bottomrule
\end{tabular}
\end{table}

The image-adaptive order improves DISTS over the static importance order by
0.002, 0.003, and 0.002 at 25\%, 50\%, and 75\% of the plane, respectively.
Its 84-bit side information costs $84/(HW)$ bpp for an image of area $HW$. The
static importance order therefore preserves most of the measured DISTS gain
without this signaling cost when the fixed scale-and-raster order is not
required.

\section{Realized Rate, Distributional Metrics, and Runtime Tables}
\label{app:rate-runtime}

\subsection{Routing quality and encoding complexity}
The reported quality results use global nearest-leaf assignment, which
preserves the endpoint obtained by searching all 4096 leaves.  We evaluated a
second assignment rule that chooses the closer child at each of the twelve
tree levels.  The latter requires 24 child comparisons per token, compared
with 4096 leaf distances for global assignment.  The comparison used a
decoder-adapted checkpoint after 80,000 updates and all 24 Kodak
images at their original resolutions in a single full-resolution forward pass.
Timing was measured on one NVIDIA RTX 4090.

\begin{table}[ht]
	\centering
	\caption{Measured routing--quality trade-off on the 24-image Kodak set.
		Routing time is measured for the 2,016 tokens in one full-resolution
		$512\times768$ image; encoder time includes analysis-transform computation
		and token assignment.  Quality metrics are reported after twelve tree
		refinements.}
\label{tab:routing}
\small
\begin{tabular}{@{}lrrrrrr@{}}
		\toprule
		Assignment & Dist./token & Route (ms) & Encode (ms) & PSNR $\uparrow$ & DISTS $\downarrow$ & Agreement \\
		\midrule
		Global nearest leaf & $4096$ & $1.99$ & $27.13$ & $22.727$ & $0.0856$ & $100.0\%$ \\
		Greedy tree & $24$ & $1.77$ & $27.01$ & $21.223$ & $0.1073$ & $15.1\%$ \\
		\bottomrule
	\end{tabular}
\end{table}

The cheaper local search reduced routing-only time by 11.2\%, but the full
encoder time changed by only 0.45\%, because the analysis transform dominates
the assignment step.  Its effect on reconstruction was depth-dependent.  At
depth six, greedy routing yielded $19.846$~dB PSNR and $0.1526$ DISTS,
compared with $19.188$~dB and $0.1654$ DISTS for global assignment.  At the
full depth, however, the greedy endpoint was 1.504~dB worse and had a DISTS
value higher by 0.0217.  We therefore retain global nearest-leaf assignment
for the reported quality measurements and treat greedy routing as a separate
complexity--quality trade-off.

\subsection{Model size and inference efficiency}
\label{app:model-efficiency}
\begin{figure}[htbp]
\centering
\includegraphics[width=0.62\linewidth]{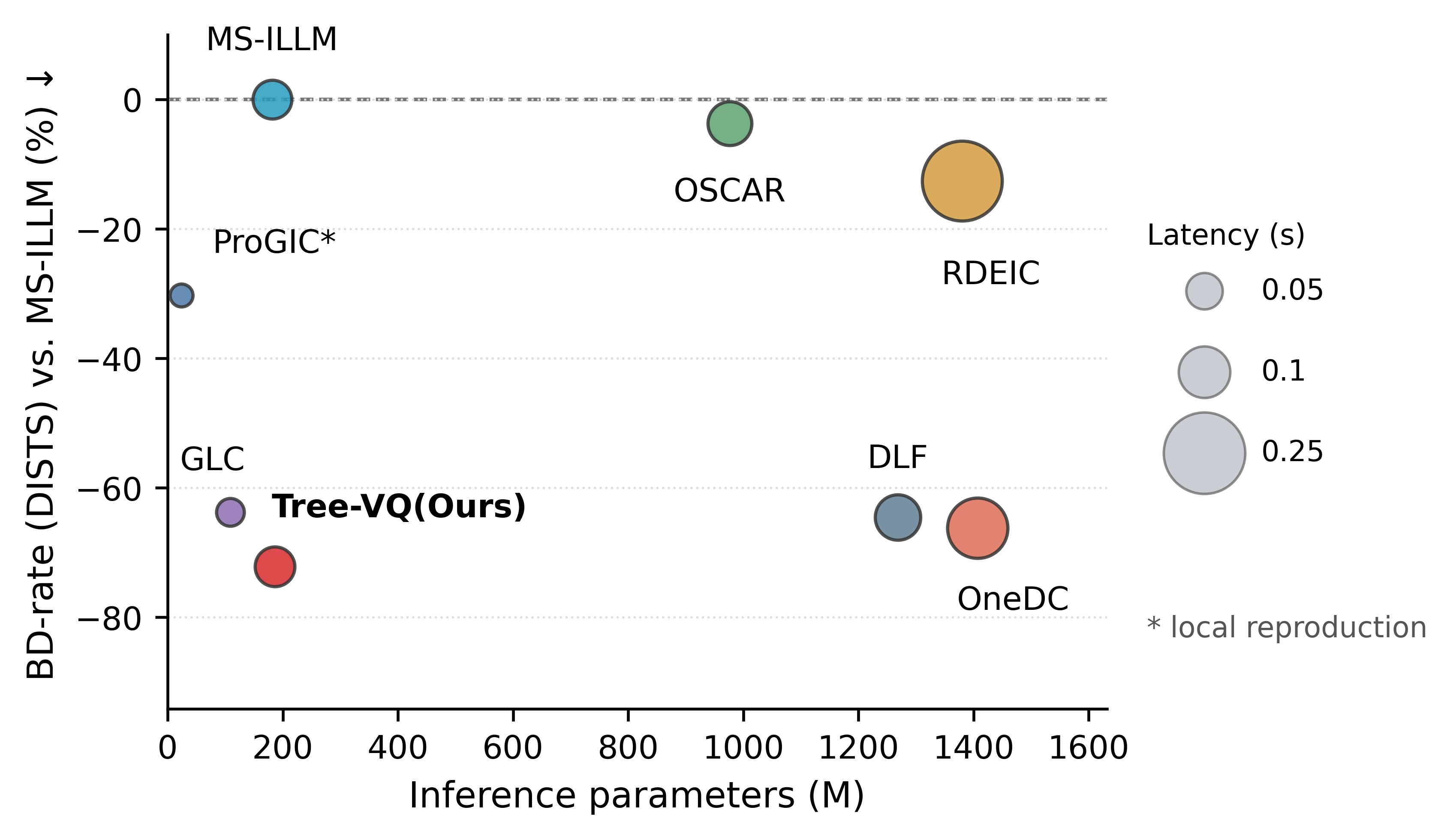}
\captionsetup{font=footnotesize,skip=2pt}
\caption{Model efficiency on DIV2K. Parameter count versus DISTS BD-rate
(lower is better); bubble area denotes latency. ProGIC* is our local
reproduction.}
\label{fig:model-efficiency}
\end{figure}

Tree-VQ uses about 190 million parameters. On Kodak, it achieves about
$52.1\%$ DISTS BD-rate saving relative to the points reported in the ProGIC
paper (Appendix~\ref{app:baseline-provenance}). Figure~\ref{fig:model-efficiency}
reports a separate model-efficiency comparison on DIV2K: GLC is smaller but weaker,
while ProGIC$^*$ is compact with a smaller gain. The figure compares parameter
count, DISTS BD-rate, and latency. All timings use the same RTX 4090.
Table~\ref{tab:routing} reports the separate routing and full-encoder times
for global nearest-leaf and greedy-tree assignment.

\section{MSE Rate--Distortion Comparison}
\label{app:mse-rd}
\begin{figure}[htbp]
\centering
\includegraphics[width=\textwidth]{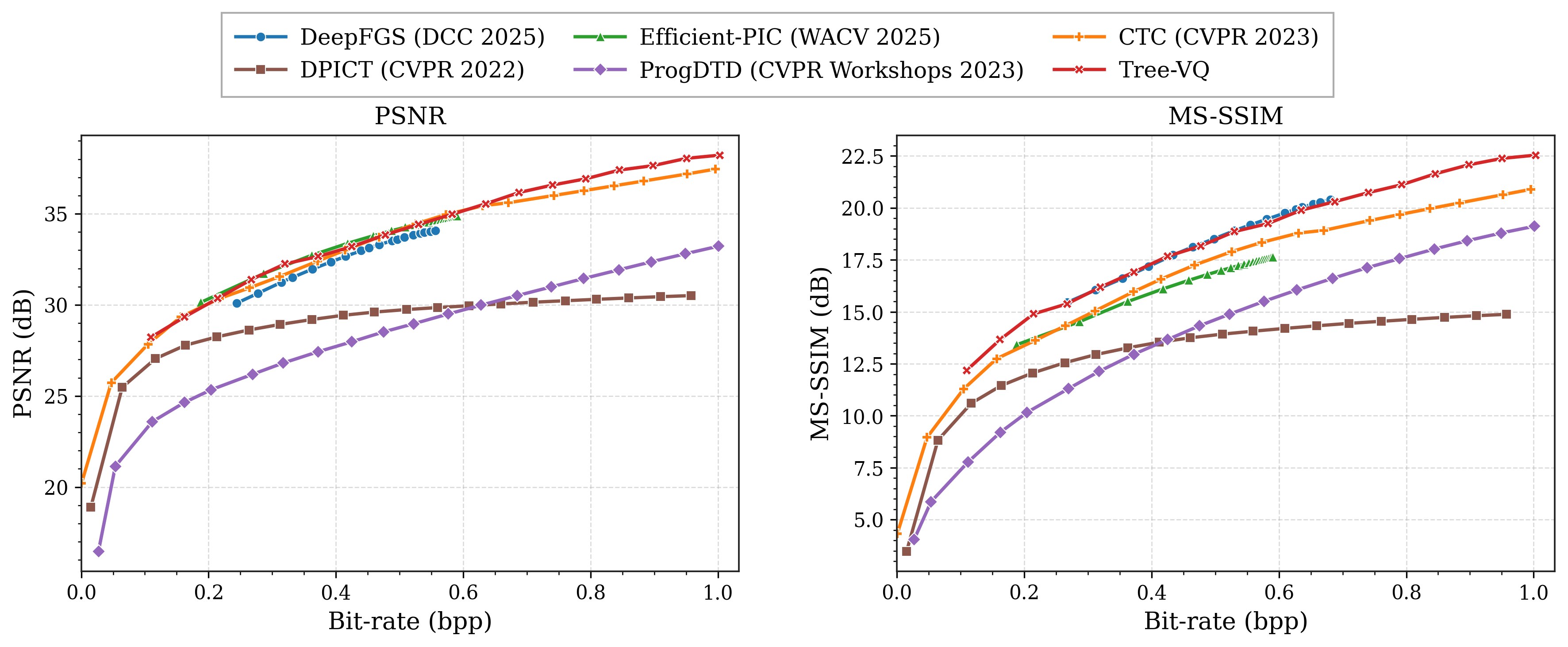}
\caption{MSE rate--distortion comparison with progressive
distortion-oriented codecs. The Tree-VQ variant retains the hierarchical
codebook, adds latent maps at $4\times$ and $8\times$ downsampling, and
uses an MSE-only objective without perceptual losses or GAN/adversarial
generative components. Higher PSNR and MS-SSIM are better.}
\label{fig:rd-mse}
\end{figure}

Figure~\ref{fig:rd-mse} complements the perceptual comparison in
Fig.~\ref{fig:perception} with a distortion-oriented evaluation. The main
comparison targets low-rate perceptual codecs, whereas many progressive
baselines optimize MSE distortion and cover a broader bitrate range. To keep
the codebook structure fixed, we reuse the Tree-VQ hierarchy and add latent
maps at $4\times$ and $8\times$ spatial downsampling. This variant uses an
MSE-only reconstruction objective and removes perceptual losses and
GAN/adversarial generative components. The resulting curves test whether the same progressive
hierarchy remains useful beyond the low-rate perceptual setting.

\section{Additional Qualitative Examples}
\label{app:qualitative}
The following figures show additional Tree-VQ reconstructions on Kodak,
DIV2K, and CLIC2020 at the rates stated in each caption.

\begin{figure}[htbp]
	\centering
	\includegraphics[width=\linewidth]{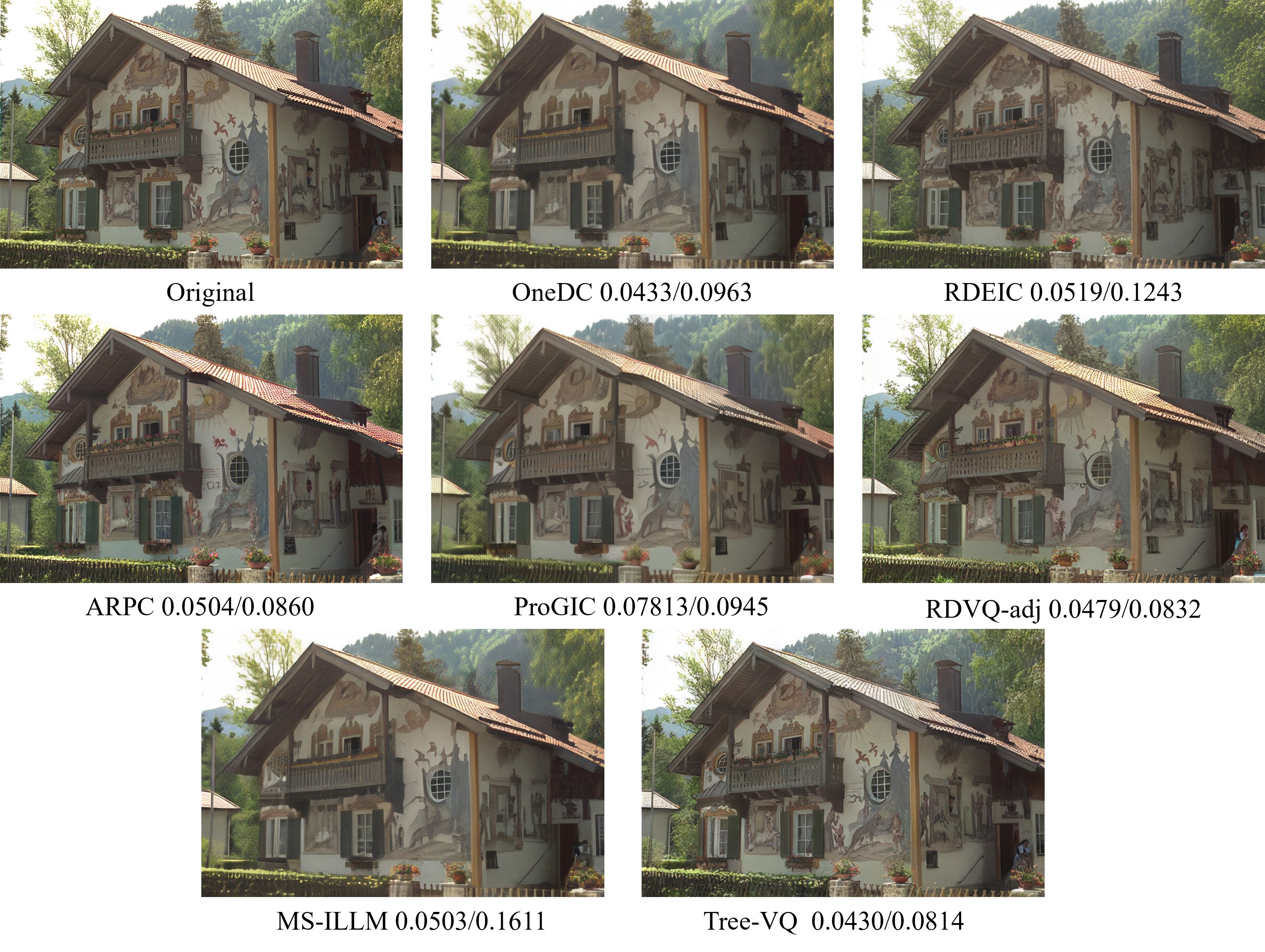}
\caption{Additional visual comparison of reconstructed images by different methods at comparable bit rates on Kodak dataset. The numbers below each image are presented as Bpp / DISTS.}
	\label{fig:recon8}
\end{figure}

\begin{figure}[htbp]
	\centering
	\includegraphics[width=\linewidth]{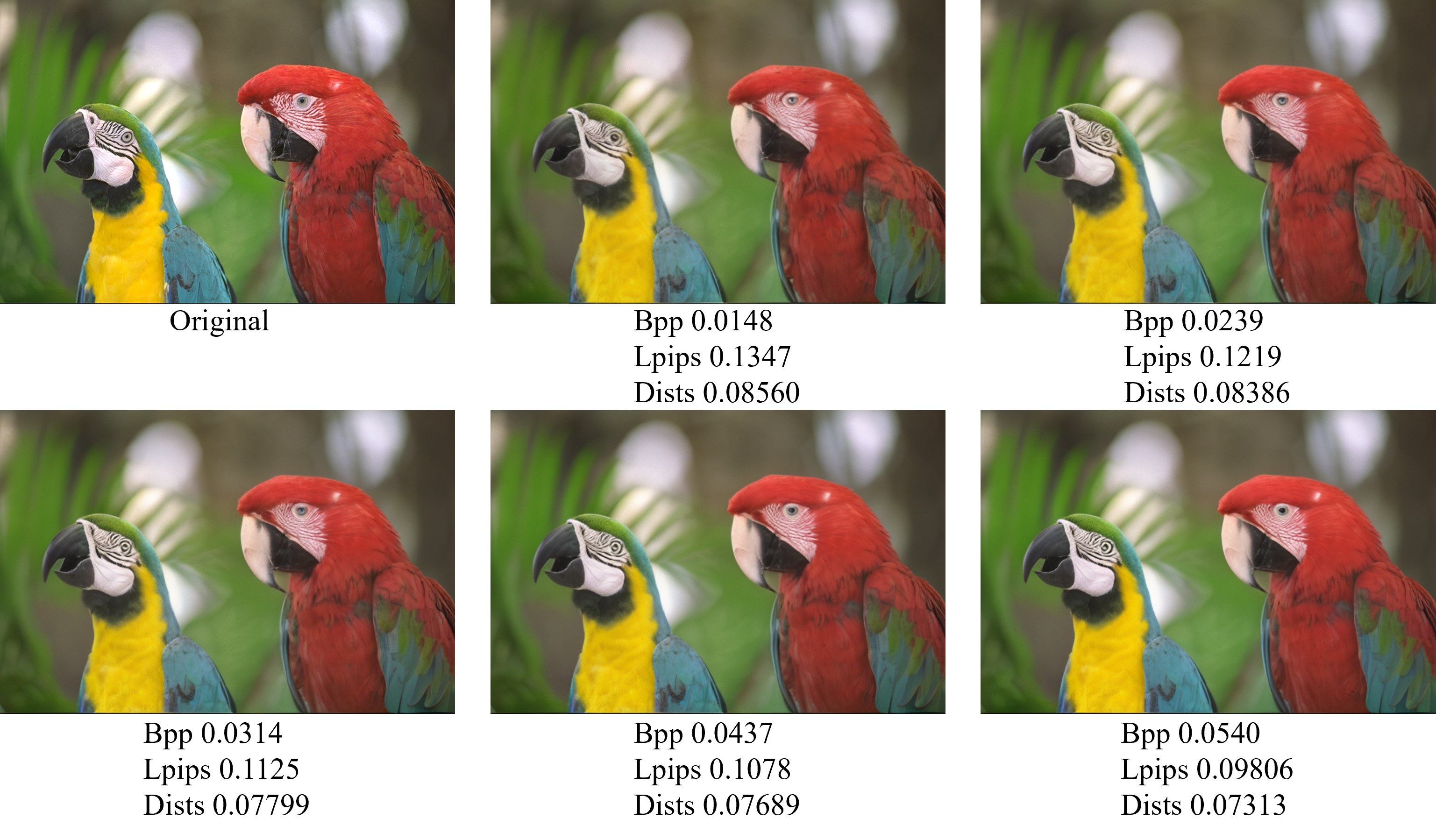}
\caption{Additional Kodak example. Read left to right and then top to bottom,
the panels show the original followed by five Tree-VQ outputs at increasing
rates spanning $0.0148$--$0.0540$ bpp.}
	\label{fig:recon5}
\end{figure}

\begin{figure}[htbp]
	\centering
	\includegraphics[width=0.65\linewidth]{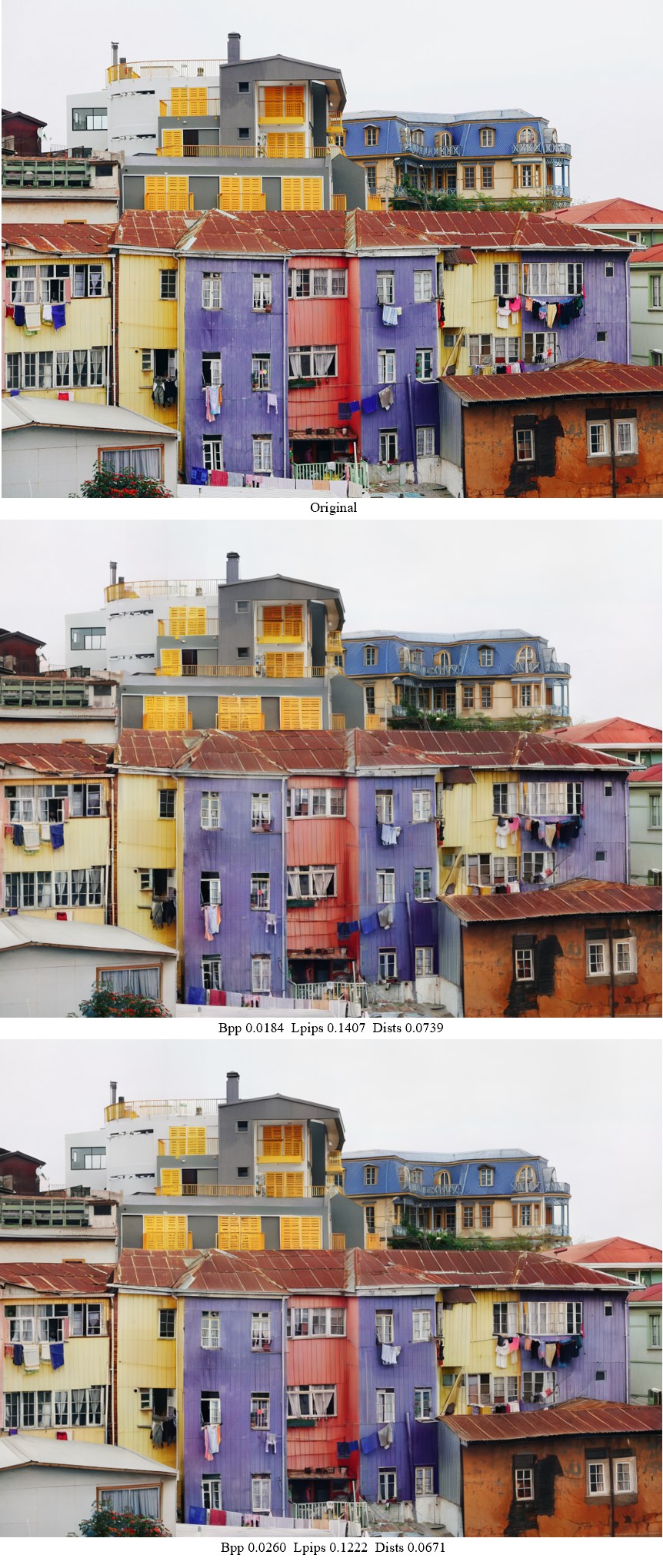}
\caption{Additional DIV2K example. From top to bottom, the panels show the
original architectural scene and Tree-VQ outputs at $0.0184$ and $0.0260$
bpp.}
	\label{fig:recon6}
\end{figure}

\begin{figure}[htbp]
	\centering
	\includegraphics[width=0.65\linewidth]{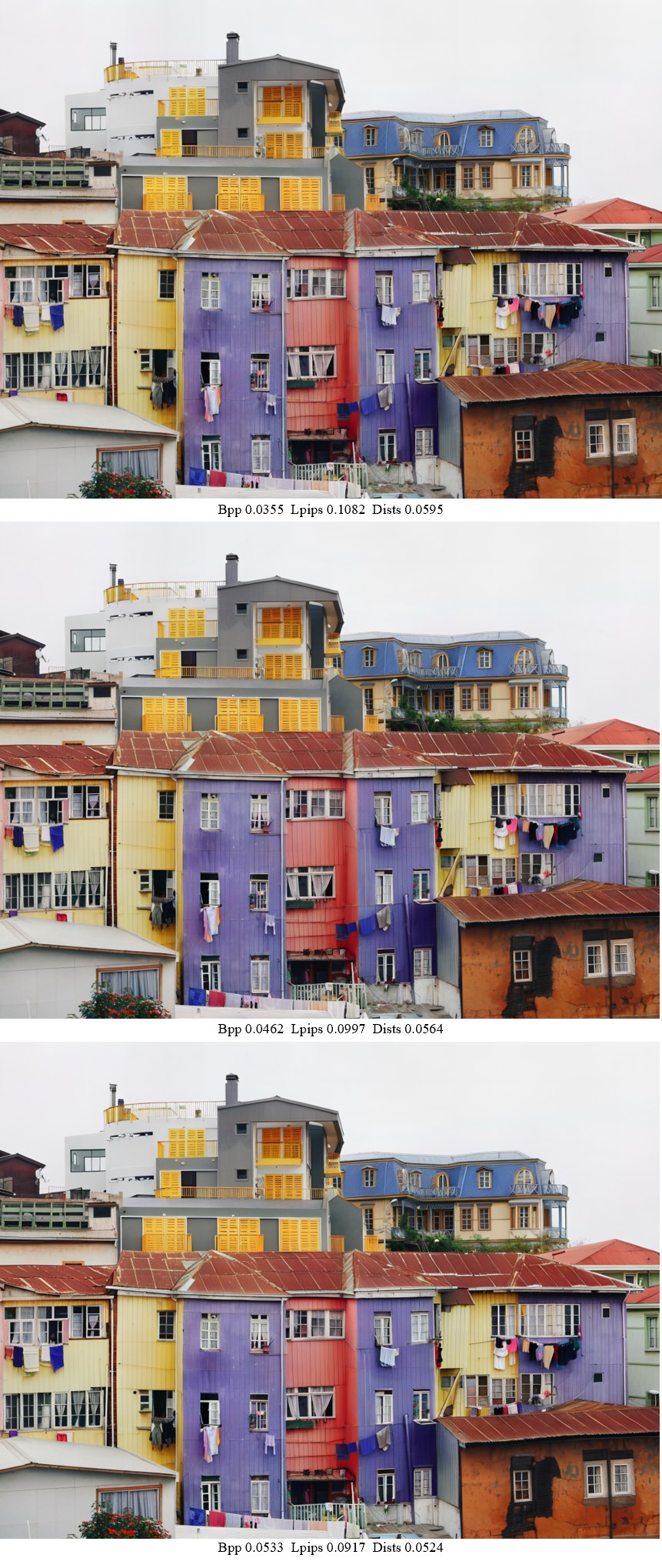}
\caption{Second DIV2K example. The three panels show Tree-VQ outputs at
$0.0355$, $0.0462$, and $0.0533$ bpp in top-to-bottom order.}
	\label{fig:recon7}
\end{figure}

\begin{figure}[htbp]
	\centering
	\includegraphics[width=0.65\linewidth]{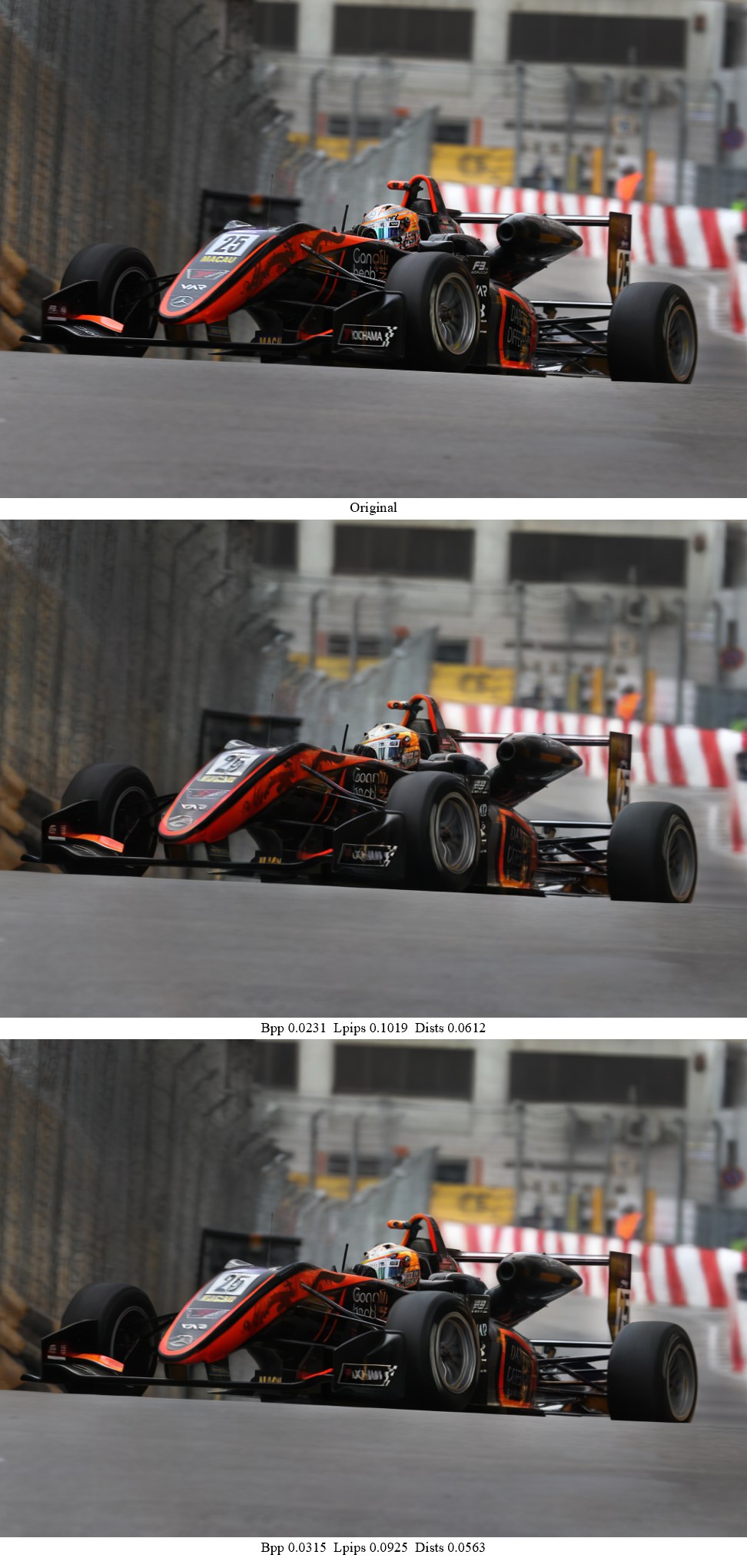}
\caption{Additional CLIC2020 example. From top to bottom, the panels show the
original racing-car image and Tree-VQ outputs at $0.0231$ and $0.0315$ bpp.}
	\label{fig:recon3}
\end{figure}

\begin{figure}[htbp]
	\centering
	\includegraphics[width=0.65\linewidth]{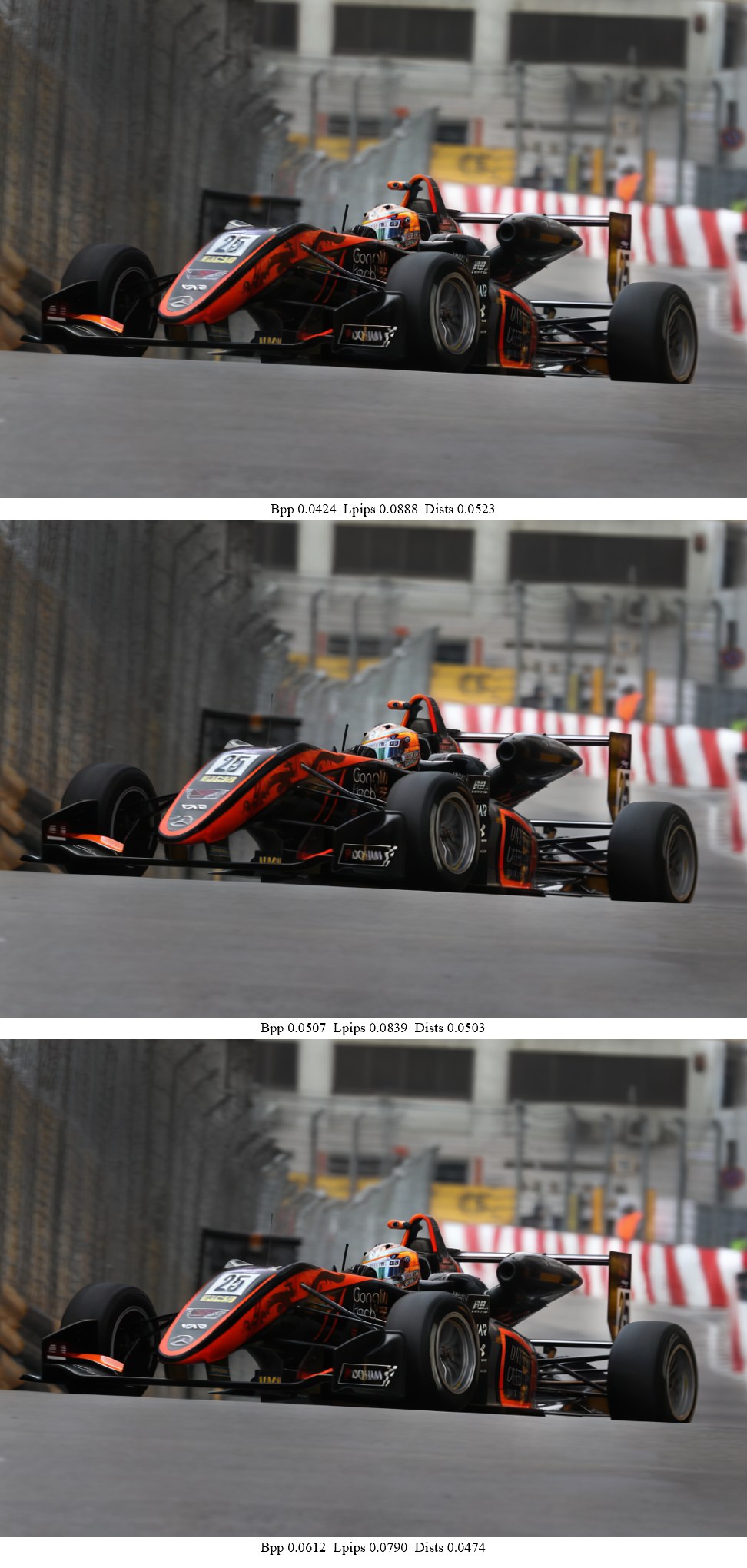}
\caption{Second CLIC2020 example. The three panels show Tree-VQ outputs at
$0.0424$, $0.0507$, and $0.0612$ bpp in top-to-bottom order.}
	\label{fig:recon4}
\end{figure}

\end{document}